\documentclass{article} 
\usepackage{iclr2025_conference,times}

\usepackage{amsmath,amsfonts,bm}

\def\eqref#1{equation~\ref{#1}}

\def\1{\bm{1}}

\DeclareMathAlphabet{\mathsfit}{\encodingdefault}{\sfdefault}{m}{sl}
\SetMathAlphabet{\mathsfit}{bold}{\encodingdefault}{\sfdefault}{bx}{n}

\usepackage{url}

\usepackage{marvosym}
\usepackage[utf8]{inputenc} 
\usepackage[T1]{fontenc}    
\usepackage{booktabs}       
\usepackage{amsfonts}       
\usepackage{nicefrac}       
\usepackage{microtype}      
\usepackage{xcolor}         

\usepackage{graphicx}
\usepackage{tabularx}       
\usepackage{comment}
\usepackage{soul}
\usepackage{amsmath}
\usepackage{amssymb}
\usepackage{float}
\usepackage{multirow}
\usepackage{xcolor}
\usepackage{lipsum}
\usepackage{footmisc}

\usepackage[colorlinks,linkcolor=red,anchorcolor=blue,citecolor=green]{hyperref}
\definecolor{green}{RGB}{0,150,10}
\definecolor{blue}{RGB}{0,148,181}
\definecolor{orange}{RGB}{194,153,107}

\usepackage[capitalize]{cleveref}
\crefname{section}{Sec.}{Secs.}
\Crefname{section}{Section}{Sections}
\Crefname{table}{Table}{Tables}
\crefname{table}{Tab.}{Tabs.}
\usepackage{colortbl}
\usepackage{textcomp}
\renewcommand{\paragraph}[1]{\vspace{0.1em}\noindent\textbf{#1}}

\usepackage{ulem} 

\title{GeoX: Geometric Problem Solving Through Unified Formalized Vision-Language Pre-training}

\author{Renqiu Xia$^{1,2,*}$, Mingsheng Li$^{2,3,*}$, Hancheng Ye$^2$, Wenjie Wu$^1$, Hongbin Zhou$^2$, \\ \bf{Jiakang Yuan$^{3}$}, \bf{Tianshuo Peng}$^{2,4}$, Xinyu Cai$^2$, Xiangchao Yan$^2$, Bin Wang$^2$, Conghui He$^2$, \\ \bf{Botian Shi$^{2}$}, \bf{Tao Chen}$^{3,}$\textsuperscript{\Letter}, Junchi Yan$^{1}$, \bf{Bo Zhang}$^{2,\ddagger,}$\textsuperscript{\Letter} \\[1.5mm]
$^1$ Shanghai Jiao Tong University, $^2$ Shanghai Artificial Intelligence Laboratory \\ $^3$ Fudan University, $^4$ MMLab, The Chinese University of Hong Kong  \\[1mm]
\normalsize * Equal Contribution, {\normalsize \Letter \ Corresponding Authors}, \normalsize $\ddagger$~Project Leader \\
}

\iclrfinalcopy 

\begin{document}

\maketitle
\fancyhead[L]{preprint}

\begin{abstract}

Despite their proficiency in general tasks, Multi-modal Large Language Models (MLLMs) struggle with automatic Geometry Problem Solving (GPS), which demands understanding diagrams, interpreting symbols, and performing complex reasoning. This limitation arises from their pre-training on \textit{natural images and texts}, along with the \textit{lack of automated verification} in the problem-solving process. Besides, current geometric specialists are limited by their \textit{task-specific designs}, making them less effective for broader geometric problems. To this end, we present GeoX, a multi-modal large model focusing on geometric understanding and reasoning tasks. Given the significant differences between geometric diagram-symbol and natural image-text, we introduce \textbf{unimodal pre-training} to develop a diagram encoder and symbol decoder, enhancing the understanding of geometric images and corpora. Furthermore, we introduce \textbf{geometry-language alignment}, an effective pre-training paradigm that bridges the modality gap between unimodal geometric experts. We propose a \textbf{Generator-And-Sampler Transformer} (GS-Former) to generate discriminative queries and eliminate uninformative representations from unevenly distributed geometric signals. Finally, GeoX benefits from visual instruction tuning, empowering it to take geometric images and questions as input and generate verifiable solutions. Experiments show that GeoX outperforms both generalists and geometric specialists on publicly recognized benchmarks, such as GeoQA, UniGeo, Geometry3K, and PGPS9k. Our code is available at \url{https://github.com/Alpha-Innovator/GeoX}

\end{abstract}

\section{Introduction}
\label{sec:introd}
\vspace{-0.2cm}

\begin{figure}[tb!]
    \vspace{-10pt}
    \centering
    \includegraphics[width=0.98\linewidth]{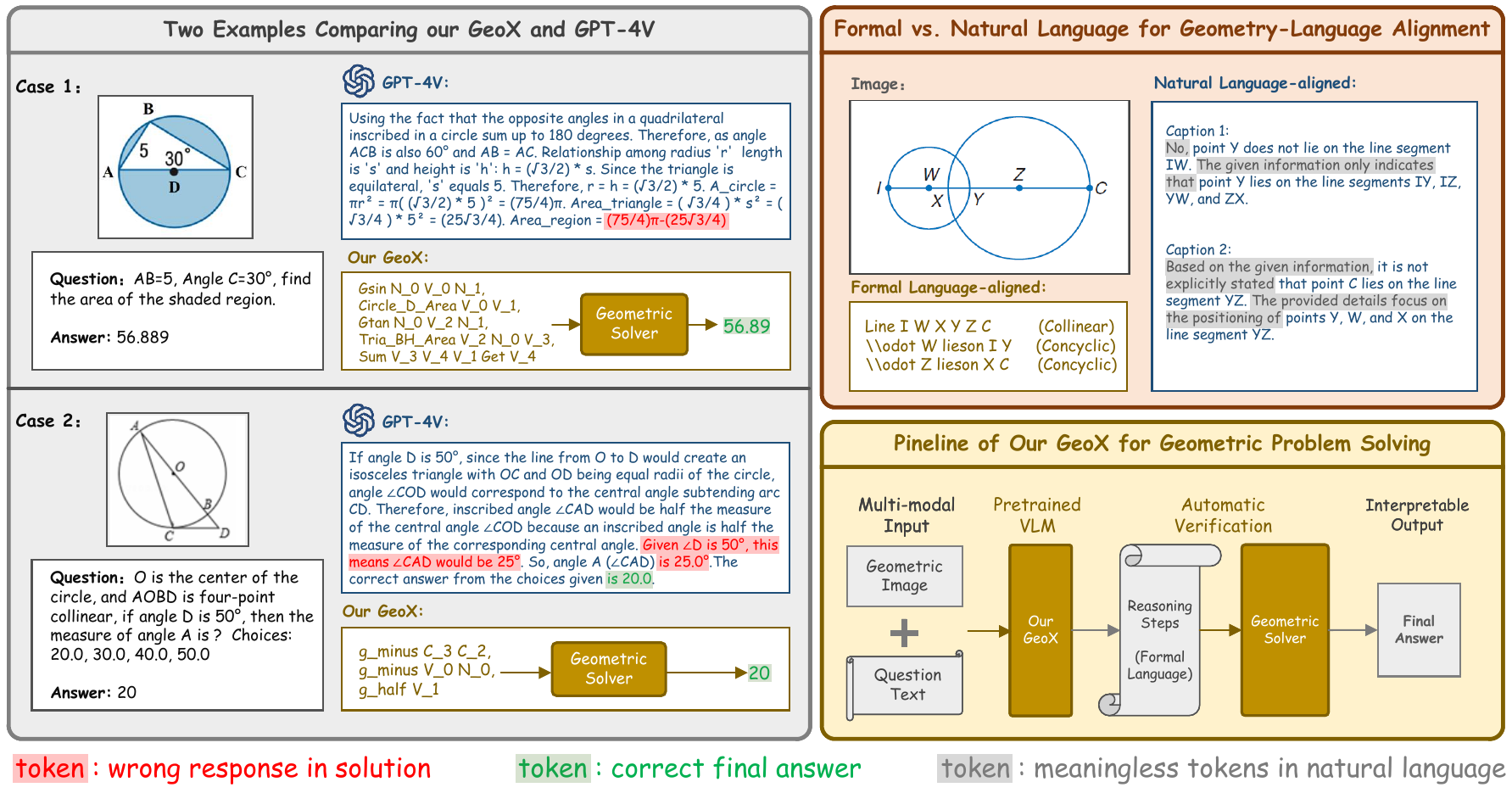}
    \vspace{-6pt}
    \caption{Highlights of GeoX: 1) 
    Comparison between GPT-4V~\citep{openai2023gpt4v} and GeoX: 
    GPT-4V often fails to provide the expected results or solving approaches. 
    Besides, verifying GPT-4V’s solutions is labor-intensive, requiring expert knowledge and step-by-step analysis.
    2) 
    Comparison between formal and natural (informal) language: 
    Unlike existing works~\citep{g-llava/gao2023g,mavis/zhang2024mavis} that use natural language, we advocate for formal language due to its effectiveness and verifiability, making it more suitable for geometric tasks.
    3) GeoX solves geometric tasks in a unified format by taking geometric images and questions as input, generating verifiable program sequences, and performing solving with a solver.
    }
    \label{fig:teaser}
    \vspace{-5pt}
\end{figure}





Large Language Models (LLMs)~\citep{llama/touvron2023llama,chatgpt/ouyang2022training} and their multi-modal extensions (MLLMs) \citep{llava/liu2024visual,chen2024far,openai2023gpt4v,2024claude} have demonstrated exceptional abilities to effectively handle a wide range of general domain tasks, such as cross-modal retrieval~\citep{wiki-llava/caffagni2024wiki,zhang2023uni3d,retrieval/wang2024multimodal,xia2024docgenome,xia2024chartx}, visual question answering~\citep{vqa1/wu2024v,ll3da/chen2024ll3da}, and summarization~\citep{llm_caption1/bianco2023improving,llm_caption2/rotstein2023fusecap}. 
With the increasing focus on Artificial General Intelligence (AGI), both LLMs and MLLMs are making inroads into specialized domains such as mathematics reasoning~\citep{mathprompter/imani2023mathprompter,wang2024unimernet,wang2024mineru}, demonstrating promising performance improvements.

Plane geometry is a pivotal and unique branch of mathematics that requires the integration of multi-modal data as well as knowledge from different scientific fields, such as theorem proving~\citep{alphageometry/trinh2024solving} and algebraic computation~\citep{algebraic/faulstich2024coupled}. 
However, developing AI systems to automatically solve geometry problems is challenging due to the inherent complexity of both visual and language modalities.
Previous works \citep{geodrl/peng2023geodrl,e-gps/wu2024gps} rely on additional detection models and make decisions based on manually crafted rules, but are often criticized for their \textit{complexity} \citep{lans/zhang2023lans}.
On the other hand, NGS~\citep{geoqa_ngs/chen2021geoqa}, Geoformer~\citep{unigeo_geoformer/chen2022unigeo}, and PGPSNet~\citep{pgpsnet/zhang2023multi} focus on predicting program sequences, yet they often suffer from \textit{poor adaptability} due to their \textit{task-specialized model designs} and \textit{limited ability} in modeling complex geometric diagrams and problems.

Although MLLMs~\citep{math-llava/shi2024math,mathvista/lumathvista} have made significant progress in multi-modal mathematical reasoning, their performance still lags behind that of specialized geometry models.
Notably, they sometimes exhibit an interesting phenomenon where they generate a correct answer accompanied by an \textit{incorrect solution process or solving approach}, as shown in \cref{fig:teaser}. 
Besides, we observe that using natural language to describe geometric diagrams introduces a significant amount of redundant information. 
In contrast, formal descriptions are more concise and clear, providing necessary information about symbols, shapes, numbers, and their relationships, making them better suited for geometric multimodal pre-training.
To this end, we argue that effectively leveraging multimodal information from both visual and textual sources through formalized pre-training is meaningful in mitigating the challenges that MLLMs face when solving geometric problems.

However, combining visual and symbolic information for pre-training to boost the ability of GPS is challenging, due to the following two reasons: \textbf{1) Large Domain Gap for Geometric Understanding.} Prior works~\citep{g-llava/gao2023g,math-llava/shi2024math} adopt a frozen CLIP ViT~\citep{clip/radford2021learning} as the diagram encoder, which is trained on natural images rich in colors and textures. However, geometric diagrams are usually monochrome, composed of elements like lines, shapes, and symbols, exhibiting a significant domain discrepancy.
\textbf{2) Uninformative Representations for Geometric Reasoning.} In geometric images, useful information is concentrated in specific areas, while other regions are uninformative and considered noise. The inability to handle this uneven distribution of geometric information leads to suboptimal performance.

To address these challenges, we propose GeoX, a geometry-centric large model that can comprehend geometric problems and solve geometry tasks in a unified formulation. 
To this end, we propose a formalized training scheme that consists of three progressive stages: unimodal pre-training, formalized geometry-language alignment, and visual instruction tuning.
%
%
In the first stage, as introduced in Sec.~\ref{sec:pre-training}, we focus on integrating a visual encoder with prior knowledge of geometry by masked auto-encoding. At the same time, we train a geometric decoder in an auto-regressive manner to enhance its comprehension of the geometry language, which is interleaved with numbers, symbols, and words.
Furthermore, solving geometric problems often requires not just recognizing shapes or symbols but also reasoning their interactions and implications. 
Thus, as described in Sec.~\ref{sec:alignment}, we introduce geometry-language alignment, which utilizes formalized descriptions instead of natural language captions, offering a new perspective to effectively align geometry-semantic features.
We present a Generator-and-Sampler Transformer (GS-Former), capable of generating geometry content-aware queries and removing uninformative representations under the guidance of semantic learning.
In Sec.~\ref{sec:tuning}, to enable GeoX to generate solutions based on the input geometric problem and image, we adopt end-to-end visual instruction tuning to obtain the ultimate model. %
Furthermore, in \cref{app:sec:thm}, we theoretically explain why the proposed formalized pre-training is more effective in GPS tasks.

In Sec.~\ref{sec:exp}, we conduct extensive experiments on four widely recognized benchmarks to evaluate GeoX's ability in reasoning complex and diverse geometric problems, where our approach achieves state-of-the-art results.
Insightful analyses and ablation experiments are performed to further validate the effectiveness of our method.

Our contributions can be summarized as follows:
\begin{itemize}
    \item Our study reveals the large potential of formalized visual-language pre-training in enhancing geometric problem-solving abilities. To enable the formalized pre-training, we propose GeoX, aiming to build geometric generalist models by modeling geometric tasks into a unified formulation.
    \item We analyze the unique challenges in the field of geometry problem solving and propose GS-Former, which effectively bridges the modality gap between geometric diagrams and formalized language.
    \item Compared with previous generalist and specialized models, our GeoX achieves competitive performance on GeoQA, UniGeo, Geometry3K, and PGPS9K, further demonstrating GeoX as a strong baseline model for solving geometric problems and motivating future research.
\end{itemize}

\vspace{-0.1cm}
\section{Related Works}
\vspace{-0.2cm}


\paragraph{Multi-modal Large Language Models.} The past year has witnessed the notable success of Large Language Models (LLMs) families~\citep{chatgpt/ouyang2022training,llama/touvron2023llama,llama2/touvron2023llama,internlm/team2023internlm}, showcasing near-human performance across diverse tasks.
Concurrently, researchers have made significant efforts to extend the abilities of LLMs in handling visual-related tasks, contributing to the flourishing of Multimodal Large Language Models (MLLMs)~\citep{qwen-vl/bai2023qwen, gpt-4v/achiam2023gpt, reid2024gemini}.
MLLMs typically adopt a cross-modal projector as the bridge to reconcile the modality gap between visual encoder and LLM, such as Q-former~\citep{blip-2/li2023blip} or linear layers~\citep{llava/liu2024visual}.
Although MLLMs have demonstrated impressive performance in conventional vision-language tasks~\citep{one-llm/han2024onellm,structchart/xia2023structchart,m3dbench/li2023m3dbench}, they yield unsatisfactory results when addressing multimodal mathematical problems involving geometric diagrams and symbols.
Besides, G-LLaVA~\citep{g-llava/gao2023g} and MAVIS~\citep{mavis/zhang2024mavis} train LLM on the constructed geometry datasets with descriptions in natural language form. Recently, Chimera~\citep{peng2024chimera} proposes using the general-expert collaboration masking method to effectively integrate expert knowledge into a general MLLMs.
However, as illustrated in \cref{fig:teaser}, these works face two issues: \textit{1) unable to provide the answer as required}, and \textit{2) incorrect solving steps that still result in correct answers}.
Furthermore, verifying the solving process of MLLMs is extremely costly since it requires human experts from geometric knowledge and a step-by-step examination.
To this end, we propose GeoX, which solves geometric tasks in a unified formulation and predicts verifiable solutions.

\paragraph{Geometry Problem Solving (GPS)} is a long-standing yet challenging task in mathematics, requiring models with the ability to understand geometric elements and reason with logic.
Existing automatic systems for GPS fall into two categories: rule-based approaches and neural approaches. 
Rule-based approaches~\citep{geos/seo2015solving,rule1/sachan2017learning,inter-gps/lu2021inter,geodrl/peng2023geodrl,e-gps/wu2024gps} rely on external tools like OCR to parse diagrams into texts, which are then used for logical reasoning based on path search and condition matching.
Although these methods have shown satisfactory performance in GPS, they are \textit{heavily dependent on manually crafted rules}, making them difficult to generalize to diverse geometry scenarios.
Neural approaches use networks to predict solving steps via program sequences, which are then executed by the solver.
For example, NGS~\citep{geoqa_ngs/chen2021geoqa} and Geoformer~\citep{unigeo_geoformer/chen2022unigeo} introduce auxiliary self-supervised tasks to refine diagram representations, with experiments on GeoQA~\citep{geoqa_ngs/chen2021geoqa} and UniGeo~\citep{unigeo_geoformer/chen2022unigeo} demonstrating the effectiveness of their methods. 
Other methods, such as PGPSNet~\citep{pgpsnet/zhang2023multi} and LANS~\citep{lans/zhang2023lans}, integrate structural and semantic clauses into solving process and utilize specially designed decoders to achieve better performance on both Geometry3K~\citep{inter-gps/lu2021inter} and PGPS9K~\citep{pgpsnet/zhang2023multi}.
While these geometry specialists have shown impressive performance, their \textit{uniquely designed models} for specialized datasets limit their ability to solve broader geometric tasks. 
In contrast, we introduce the unified formalized vision-language pre-training for general geometric tasks, achieving superior results across diverse benchmarks compared to previous methods on GPS.

\vspace{-0.2cm}
\section{Formalized Vision-Language Pre-training}
\vspace{-0.2cm}

\subsection{Method Overview} 
\begin{figure}[tb!]
    \vspace{-8pt}
    \centering
    \includegraphics[width=0.96\linewidth]{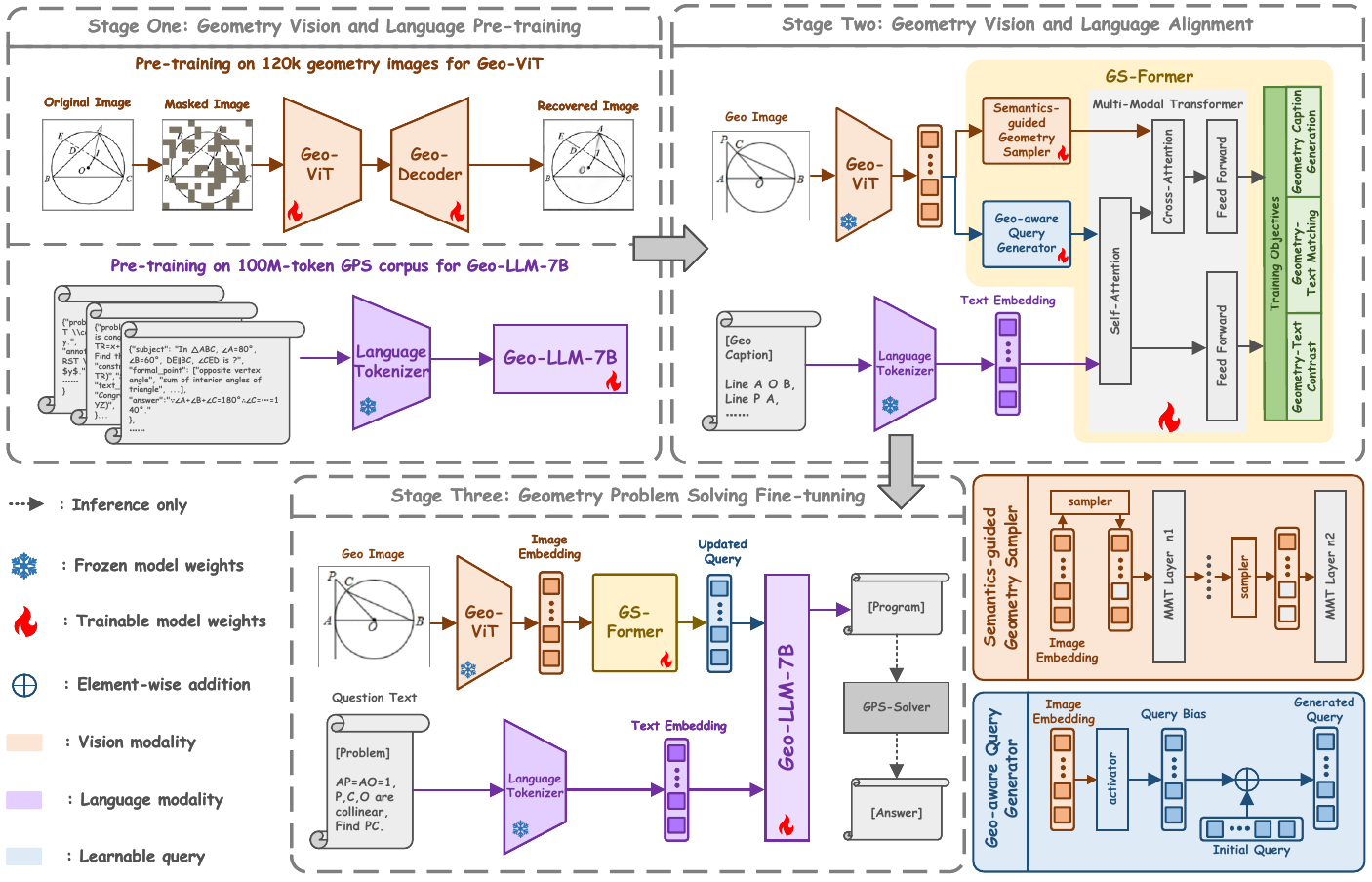}
    \vspace{-6pt}
    \caption{Overview of GeoX for training. We present a versatile method for automatic geometric problem solving through unified formalized vision-language pre-training, which comprises three progressive stages.
    }
    \label{fig:framework}
    \vspace{-12pt}
\end{figure}

\vspace{-0.2cm}
To tackle complicated plane geometry problems, we introduce GeoX, adopting a formalized pre-training scheme consisting of three progressive stages, as illustrated in~\cref{fig:framework}.

\noindent \textbf{Unimodal Pre-training.} Vanilla generalist models~\citep{openai2023gpt4v,2024claude,team2023gemini,qwen-vl/bai2023qwen, peng2024chimera,chen2024far} have poor representation capacity in the geometric domain, due to the significant gaps between non-formalized data (\textit{e.g.}, informal text descriptions and natural images) and formalized data (\textit{e.g.}, formal geometric symbols and scientific images). As a result, we propose unimodal pre-training in~\cref{sec:pre-training}, aiming to enhance the GeoX's ability to understand geometric diagrams and symbols.

\noindent \textbf{Geometry-Language Alignment.} To facilitate the aforementioned pre-trained unimodal models for performing cross-modal alignment, we propose an effective Generator-and-Sampler Transformer (GS-Former), which is trained using pairs of geometric diagrams and formal language descriptions, as detailed in~\cref{sec:alignment}.

\noindent \textbf{End-to-end Instruction Tuning.} After geometry-language alignment, the ultimate model is required to generate solutions based on the given geometric problems and images. To this end, we tune GeoX in an end-to-end visual instruction tuning manner (as introduced in~\cref{sec:tuning}), boosting its capacity to comprehend geometric problems and generate formal solution programs.

During the inference phase, the solution generated by GeoX is fed into the 
symbolic solver~\citep{geoqa_ngs/chen2021geoqa,pgpsnet/zhang2023multi}, which performs step-by-step operations to predict the final answer.

\subsection{Unimodal Pre-training}
\label{sec:pre-training}
\vspace{-0.2cm}



\paragraph{Geometry Encoder.} To mitigate the deficiencies of the existing visual encoders in comprehending geometric images, we collect more than 120K diagrams\footnote[1]{\url{https://huggingface.co/datasets/U4R/GeoX-data}\label{fn:1}} from the web and electronic textbooks to equip ViT with prior knowledge of geometry, abbreviated as Geo-ViT. Similar to~\citet{mae/he2022masked}, we tune the vision encoder-decoder using the masked auto-encoding scheme, where some patches are masked and the remaining subset is fed into the visual encoder, with the original image subsequently reconstructed by a lightweight decoder. In the next stages, we only utilize the pre-trained encoder to represent geometric diagrams.

%
    
\paragraph{Symbol Decoder.} Considering the capability of LLMs to follow users' instructions and handle different tasks, we utilize the decoder-only LLM as our symbol decoder to generate solutions.
However, LLMs~\citep{gpt3/brown2020language,llama2/touvron2023llama} are typically trained on general text, which lacks the specialized learning for geometry.
To this end, we build a 100M-token geometric corpus based on the existing datasets~\citep{geoqa_ngs/chen2021geoqa,inter-gps/lu2021inter,g-llava/gao2023g, pgpsnet/zhang2023multi, unigeo_geoformer/chen2022unigeo, dpe-gps/cao2022augmented}, containing a wide range of geometric problems, symbols, theorems, and so on. More details can be found in \cref{app:sec:corpus}.
%
%
We choose LLEMMA-7B~\citep{llemma/azerbayev2023llemma} as the base model, an open-source language model for mathematics pre-trained on Proof-Pile-2~\citep{llemma/azerbayev2023llemma}, and further fine-tune it on the geometric corpus using a standard auto-regressive language modeling objective, resulting in Geo-LLM-7B.

%


\subsection{Geometry-language Alignment}
\label{sec:alignment}
\vspace{-0.2cm}

\subsubsection{Data Engine} 
\vspace{-0.2cm}


While recent datasets~\citep{g-llava/gao2023g,mavis/zhang2024mavis} have made strides in captioning geometric images using natural language, they often result in redundant information, as depicted in \cref{fig:teaser}. In contrast, our approach emphasizes the use of formal descriptions to encapsulate the spatial structural information within geometric images. This information is implicitly represented, not explicitly stated in the problem text.
Our curated dataset\footref{fn:1} focuses on capturing the essence of geometric imagery by detailing the relationships between the most fundamental elements (points) without explicitly annotating higher-level constructs such as squares or triangles, which can either be inferred from the relationships we describe or are directly provided in the problem text.

Our formalized diagram-caption dataset delves into the spatial relationships at a granular level, starting with the basic building blocks of geometric images. We identify and describe the relative positions and connections between points, ensuring that the spatial relationships are accurately represented. These relationships are categorized into two primary types: \textbf{1)} Collinear Relationship (\textit{e.g.}, \texttt{line A B C} signifies that points A, B, and C are on the same line) and \textbf{2)} Concyclic Relationship (\textit{e.g.}, \texttt{$\backslash\backslash$odot O lieson A B C} denotes that points A, B, and C are on the same circle with center O). The dataset encompasses 6232 geometric images sourced from the internet, meticulously annotated by a team of 10 experts over a period of 200 hours. Moreover, we provide concrete examples along with comprehensive explanations of formalized diagram-caption pairs in \cref{app:sec:formal_data}.


\vspace{-0.2cm}
\subsubsection{Generator-and-Sampler Transformer}
\vspace{-0.2cm}

With the formalized geometry-language dataset, GeoX learns a unified representation space for geometry and formalized language through the Generator-and-Sampler Transformer (GS-Former), which includes a Geo-aware Query Generator (GQG) and a Multi-Modal Transformer.
%

%

\paragraph{Geo-aware Query Generator.} Both Resampler~\citep{flamingo/alayrac2022flamingo,mimic/li2023mimic} and Q-Former~\citep{blip-2/li2023blip,instructblip/instructblip} extract visual features using a set of static query tokens, which are randomly initialized and regarded as model parameters. 
However, these queries, which remain the same for different diagrams, fail to capture discriminative features unique to individual samples.
Thus, we introduce the Geo-aware Query Generator (GQG), which incorporates contextual information to dynamically generate queries.


To be specific, GQG utilizes visual features from the encoder and aggregates contextual information through an attention-based module and pooling operation. The contextual features then are projected and added with learnable queries \citep{blip-2/li2023blip}, which builds a connection between the learnable queries and the geometric content. 
Our empirical results demonstrate the effectiveness of GQG, resulting in improved performance.



\paragraph{Multi-Modal Transformer} comprises $N_L$ layers, each containing a self-attention block, a cross-attention block, and a feed-forward network. Queries within each layer initially interact with paired formal captions and are then fed into the cross-attention block to extract visual features.

To handle the uneven information distribution in geometric images as described in Sec.~\ref{sec:introd}, we introduce the Semantics-guided Geometry Sampler (SGS), which dynamically removes uninformative visual representations guided by vision-language alignment.
Specifically, SGS is tasked with predicting a binary mask \(M = \{m^{i}_{j} \mid i \in K, j \in N\}\), with each \(m^{i}_{j} \in \{0, 1\}\) determining whether to retain or discard visual representations.  Here, \(K\) represents the layer number and \(N\) denotes the number of patches.
This module receives the previous mask \(M^{i-1}\) and visual features as inputs, using a linear layer to obtain retention probabilities \(P^i\). 
%
To enable differentiable sampling from probabilities, we use the reparameterization~\citep{gumbel-softmax/jang2016categorical} with Gumbel-Softmax:
\begin{equation}
    M^i = M^{i-1} \odot \text{Gumbel-Softmax}(P^i),
\end{equation}
where \(\odot\) is the Hadamard product, \(i\) and \(i-1\) represents the previous stage and current stage.
A notable feature of our GS-Former is its capability to progressively drop noisy and semantically irrelevant features under the guidance of geometric language alignment.
This is achieved by initially setting all elements of the decision mask to 1, followed by inserting the SGS block at subsequent layers.
Additionally, GS-Former is initialized with weights from pre-trained BERT models~\citep{bert/kenton2019bert}, except for the SGS and cross-attention layers, which are initialized randomly.
Inspired by BLIP-2~\citep{blip-2/li2023blip}, we introduce a multimodal alignment loss $\mathcal{L}_{align}$ to optimize GS-Former, incorporating three training objectives: Geometry-Text Contrast and Geometry-Text Matching, both designed to align features between geometric diagrams and formal text, along with Geometry Caption Generation, aimed at generating formal captions based on visual information.
We further impose a sparsification term $\mathcal{L}_{spr}$ into the overall optimization objective to prevent trivial solutions where all mask values \(m^{i}_{j}\) are set to 1:
\begin{equation}
    \small
    \mathcal{L}_p = \mathcal{L}_{align} + \lambda \mathcal{L}_{spr}, \text{ where } \mathcal{L}_{spr} = \frac{1}{KN} \sum_{i \in K,j \in N} \left\lVert m^{i}_{j} \right\rVert_1 .
    \label{eq:loss_function}
\end{equation}


\subsection{End-to-end Visual Instruction Tuning}
\label{sec:tuning}
\vspace{-0.2cm}

To enable the model to handle geometry-centric tasks, we continue the training with end-to-end visual instruction tuning, directing the ultimate model to generate solutions.
As illustrated in \cref{fig:framework}, we feed the diagrams into the pre-trained Geo-ViT together with GS-Former, to obtain the semantically aligned geometry features $F_{g}$. 
Besides, we utilize a trainable projection head $\mathcal{W}$ to project $F_{g}$ into the language embedding space and obtain visual tokens $T_{g}$.
%
Geo-LLM, serving as a decoder for various geometry tasks, takes both visual tokens $T_{g}$ and instruction tokens $T_{p}$ as input, and generates solutions in an auto-regressive manner.
Our training objective is to optimize the GeoX so that the likelihood of the target sequence $S =\{s_{i, i \in [1:L]}\}$ is maximized given the visual input $T_{g}$ and instruction $T_{p}$. 

In practice, GeoX is trained using cross-entropy loss $\mathcal{L}_t$ defined as follows, which optimizes the model to predict the $l$-th token \(s_l\) given preceding token sequences $s_{i, i \in [1:l-1]}$:
\begin{equation}
    \mathcal{L}_t = -\sum_{l} \log P (s_{l} | s_{i, i \in [1:l-1]}; T_{g}; T_{p}).
\end{equation}
\vspace{-0.1cm}
\section{Experiments}
\label{sec:exp}

\subsection{Datasets, Metrics, and Implementation Details}
\vspace{-0.15cm}

\paragraph{Datasets.} To assess the effectiveness of GeoX, we conduct experiments on four widely recognized geometry benchmarks: GeoQA~\citep{geoqa_ngs/chen2021geoqa}, UniGeo~\citep{unigeo_geoformer/chen2022unigeo}, Geometry3K~\citep{inter-gps/lu2021inter}, and PGPS9K~\citep{pgpsnet/zhang2023multi}. GeoQA comprises 4,998 geometry problems sourced from Chinese middle school exams, including different types of problems, such as angles, lengths, and areas.
%
Following \citet{unimath/liang2023unimath, g-llava/gao2023g}, we use the English version to maintain linguistic consistency with other datasets.
UniGeo features 4,998 calculation problems from GeoQA and 9,543 proving problems from high school textbooks and online resources, providing a comprehensive benchmark for evaluating geometry reasoning abilities.
Both Geometry3K and PGPS9K include high-quality diagrams and detailed annotations. 
%
%

\paragraph{Metrics.} We adopt the same evaluation metrics used in previous studies to ensure fair comparability. 
Following \citet{geoqa_ngs/chen2021geoqa} and \citet{unigeo_geoformer/chen2022unigeo}, we assess the model's performance on GeoQA and UniGeo with top-1 and top-10 accuracies.
%
For evaluation on Geometry3K and PGPS9K, we apply three metrics to assess the performance of GeoX: Completion, Choice, and Top-3, as introduced in~\citet{pgpsnet/zhang2023multi}. 
To evaluate MLLMs in solving complex geometry problems, such as Qwen-VL~\citep{qwen-vl/bai2023qwen} and GPT-4V~\citep{openai2023gpt4v}, we follow LANS~\citep{lans/zhang2023lans} by utilizing Completion (which requires models to provide answers directly) and Choice (which involves selecting from given options).

\paragraph{Implementation Details.}
We optimize the diagram encoder using MAE VIT-B~\citep{mae/he2022masked} checkpoints, training it for 800 epochs with a batch size of 256 and an initial learning rate of 6.4e-5. 
We initialize the symbol decoder with LLEMMA-7B~\citep{llemma/azerbayev2023llemma} weights and train it for 5 epochs with a batch size of 32 and an initial learning rate of 1e-6.
For geometry-language alignment, we train the GS-Former for 360 epochs with a batch size of 256 and an initial learning rate of 1e-4. 
The number of queries in GS-Former is set to 8.
%
%
Additional details regarding visual instruction tuning can be found in \cref{app:sec:implementation}.
We implement GeoX using PyTorch and conduct experiments on more than eight A100 (80GB) GPUs.
%
During inference, we employ a beam search size of 10, consistent with \citet{pgpsnet/zhang2023multi} and \citet{geoqa_ngs/chen2021geoqa}.

\begin{table*}[t]
\vspace{-10pt}
    \caption{
        Comparison of various methods on the GeoQA benchmark with different accuracy metrics.
        }
    \centering
    \begin{minipage}[htbp]{0.47\textwidth}
        \centering
        \makeatletter\def\@captype{table}
        \resizebox{\textwidth}{!}{
        \begin{tabular}{ccccc}
            \toprule
            Methods &Metric& Total& Angle& Length \\ \hline
            \multicolumn{5}{l}{\textbf{\textit{Generalists}}} \\
            mPLUG-Owl2~\citep{mplug-owl2/ye2023mplug} & \multirow{9}{*}{Top-1} & 16.0 & 16.5 & 15.9 \\
            LLaVA-v1.5~\citep{llava/liu2024visual}  & & 20.7 & 20.9 & 19.8 \\
            Qwen-VL~\citep{qwen-vl/bai2023qwen} & & 24.4 & 23.7 & 24.4 \\
            GPT-4V~\citep{openai2023gpt4v} & & 43.4 & 39.3 & 49.8 \\ 
            \cline{1-1}
            \multicolumn{5}{l}{\textbf{\textit{Specialists}}} \\
            LLaVA-v1.5~\citep{llava/liu2024visual}+\textbf{Solver} & & 9.4 & 14.9 & 3.2 \\
            NGS\citep{geoqa_ngs/chen2021geoqa} & & 46.3 & - & - \\
            UniMath-T5\citep{unimath/liang2023unimath} & & 49.6 & - & - \\
            UniMath-Flan-T5\citep{unimath/liang2023unimath} & & 50.0 & - & - \\
            \rowcolor[gray]{0.9}
            GeoX (Ours)  & & \textbf{54.9} & \textbf{62.8} & \textbf{45.2} \\
            \bottomrule
        \end{tabular}
        }
    \end{minipage}
    \hspace{0.3cm} 
    \begin{minipage}[htbp]{0.47\textwidth}
        \centering
        \makeatletter\def\@captype{table}
        \resizebox{\textwidth}{!}{
        \begin{tabular}{ccccc}
            \toprule
            Methods &Metric& Total& Angle& Length \\ \hline
            \multicolumn{5}{l}{\textbf{\textit{Specialists}}} \\
            LLaVA-v1.5~\citep{llava/liu2024visual}+\textbf{Solver} & \multirow{9}{*}{Top-10} & 29.2 & 40.5 & 15.9 \\
            FiLM\citep{film/perez2018film} & & 31.7 & 34.0 & 29.7 \\
            RN\citep{rn/santoro2017simple} & & 38.0 & 42.8 & 32.5 \\
            MCAN\citep{macn/yu2019deep} & & 39.7 & 45.0 & 34.6 \\
            BERT~\citep{bert/kenton2019bert} & & 54.7 & 65.8 & 42.1 \\
            NGS\citep{geoqa_ngs/chen2021geoqa} & & 56.9 & 69.8 & 39.2 \\
            Geoformer\citep{unigeo_geoformer/chen2022unigeo} & & 60.3 & 71.5 & 49.1 \\ 
            DPE-NGS\citep{dpe-gps/cao2022augmented} & & 62.7 & 74.9 & 47.7 \\ 
            SCA-GPS\citep{sca-gps/ning2023symbolic} & & 64.1 & 74.9 & 50.1 \\
            \rowcolor[gray]{0.9}
            GeoX (Ours) & & \textbf{69.0} & \textbf{78.2} & \textbf{58.0} \\ 
            \bottomrule
        \end{tabular}
        \label{exp:geoqa}
        }
    \end{minipage}
\end{table*}

\begin{table*}[t]
\vspace{-10pt}
    \caption{
        Comparison of model performance on UniGeo for geometry calculation and proof problems.
        }
    \centering
    \begin{minipage}[htbp]{\textwidth}
        \flushright
        \makeatletter\def\@captype{table}
        \resizebox{\textwidth}{!}{
        \begin{tabular}{ccccccccccccc}

        \toprule
        \multicolumn{1}{c}{\multirow{2}{*}{Methods}} & \multicolumn{1}{c}{\multirow{2}{*}{Metric}} & \multicolumn{3}{c}{Calculation(\%)}       & &  \multicolumn{6}{c}{Proving (\%)}                                              \\ \cline{3-5} \cline{7-12}
        \multicolumn{1}{c}{}         &               &  $All \uparrow$ &  $ Angle \uparrow$ &  $Length \uparrow$ &  &$All \uparrow$  & $Par. \uparrow$  & $Tri. \uparrow$  &   $Qua. \uparrow$   &   $Con. \uparrow$  &   $Sim. \uparrow$  \\ \hline

        \multicolumn{1}{l}{\textbf{\textit{Generalists}}} \\
        mPLUG-Owl2~\citep{mplug-owl2/ye2023mplug}&\multirow{9}{*}{Top-1}& 18.7 & 18.7 & 19.1 & &- & -&- &- & - &- \\
        LLaVA-v1.5~\citep{llava/liu2024visual}  && 24.0 & 26.4 & 21.6 & & -&- & -&- & - &- \\
        Qwen-VL~\citep{qwen-vl/bai2023qwen}  && 24.4 & 24.2 & 25.4 & & -&- &- & -& - &- \\
        GPT-4V~\citep{openai2023gpt4v}  && 47.9 &45.8 & 51.6 & & -& -&- & -& - &- \\
        \cline{1-1}
        
        \multicolumn{1}{l}{\textbf{\textit{Specialists}}} \\
        LLaVA-v1.5~\citep{llava/liu2024visual}+\textbf{Solver}  &  & 16.1 & 19.2 & 13.1 & & 1.0 &0.0 &  1.1 & 0.4 & 0.2 & 3.0\\
        Geoformer~\citep{unigeo_geoformer/chen2022unigeo} & &46.8&57.8&35.0&&51.3&13.9&63.8&20.4&56.1&64.0 \\
        UniMath-T5-base~\citep{unimath/liang2023unimath}&&-&-&-&&82.9 &-&-&-&-&-   \\
        UniMath-Flan-T5-base~\citep{unimath/liang2023unimath} &&-&-&-&&83.0 &-&-&-&-&- \\
        \rowcolor[gray]{0.9} 
        GeoX (Ours)   && \textbf{54.4} & \textbf{63.1}& \textbf{43.1} && \textbf{97.8} &\textbf{77.8}&\textbf{100.0}&\textbf{95.4}&\textbf{99.5}&\textbf{99.2}\\ \hline
        
        \multicolumn{1}{l}{\textbf{\textit{Specialists}}} \\
        LLaVA-v1.5~\citep{llava/liu2024visual}+\textbf{Solver}  &\multirow{5}{*}{Top-10} & 43.0 & 51.3 & 35.3 & & 11.3&0.0 & 16.2 & 5.0 & 2.9 & 27.5\\
        BERT~\citep{bert/kenton2019bert} &   &52.0& 63.1& 39.2 &&48.1 &15.4 &48.0& 31.7& 49.5& 75.1 \\
        NGS~\citep{geoqa_ngs/chen2021geoqa} & & 51.9& 63.6 &38.8 &&47.4& 11.2& 46.9& 31.3& 48.3 &77.6 \\
        Geoformer~\citep{unigeo_geoformer/chen2022unigeo}  && 62.5 &75.5& 48.8 & &56.4& 19.4& 69.4 &20.4 &60.3 &75.0 \\ 
         \rowcolor[gray]{0.9} 
         GeoX (Ours) &&\textbf{68.6} & \textbf{76.7}&\textbf{58.3} & &\textbf{99.5} & \textbf{97.2}&\textbf{100.0} &\textbf{97.7}& \textbf{100.0} &\textbf{100.0}\\
        \bottomrule
        \end{tabular}
        }
        \label{table:unigeo}
    \end{minipage}    
\end{table*}

\begin{table*}[t]
\vspace{-10pt}
    \caption{Performance comparison on Geometry3K and PGPS9K.}
    \centering
    \begin{minipage}[htbp]{0.97\textwidth}
        \flushright
        \makeatletter\def\@captype{table}
        \resizebox{\textwidth}{!}{
        \begin{tabular}{cccccccc}
        \toprule
        \multirow{2}{*}{Methods} & \multicolumn{3}{c}{Geometry3K} && \multicolumn{3}{c}{PGPS9K} \\
        \cline{2-4} \cline{6-8}
        & $Completion\uparrow$ & $Choice\uparrow$ & $Top-3\uparrow$& & $Completion\uparrow$ & $Choice\uparrow$ & $Top-3\uparrow$ \\
        \midrule
        \multicolumn{1}{l}{\textbf{\textit{Generalists}}} \\
        mPLUG-Owl2~\citep{mplug-owl2/ye2023mplug} & 2.2 & 26.7 & - && 3.0 & 26.4 & - \\
        LLaVA-v1.5~\citep{llava/liu2024visual} & 2.9 & 22.9 & - & & 1.8 & 21.8 & - \\
        Qwen-VL~\citep{qwen-vl/bai2023qwen}& 2.5 & 27.5 & - && 1.4 & 24.7 & - \\
        GPT-4V~\citep{openai2023gpt4v} & 34.8 & 58.6 & - && 33.3 & 51.0 & - \\
        \cline{1-1}
        \multicolumn{1}{l}{\textbf{\textit{Specialists}}} \\
        LLaVA-v1.5~\citep{llava/liu2024visual}+\textbf{Solver} &  19.7 &  47.4 &  31.6  && 21.6  & 38.1  & 35.3   \\
        GeoDRL~\citep{geodrl/peng2023geodrl} & - & 68.4 & - && - & - & - \\
        NGS~\citep{geoqa_ngs/chen2021geoqa} & 35.3 & 58.8 & 62.0 && 34.1 & 46.1 & 60.9 \\
        Geoformer~\citep{unigeo_geoformer/chen2022unigeo} & 36.8 & 59.3 & 62.5 & & 35.6 & 47.3 & 62.3 \\
        %
        InterGPS~\citep{inter-gps/lu2021inter} & 44.6 & 56.9 & - && - & - & - \\
        PGPSNet~\citep{pgpsnet/zhang2023multi} & 48.1 & 70.1 & 65.7 & & 44.4 & 57.6 & 64.8 \\
        \rowcolor[gray]{0.9}
        GeoX (Ours) &\textbf{58.6} & \textbf{72.5} & \textbf{69.4}   && \textbf{52.7} & \textbf{63.3} & \textbf{65.4} \\ 
        \bottomrule
        \end{tabular}
        }
        \label{table:3k9k}
        \vspace{-15pt}
    \end{minipage}    
\end{table*}

\subsection{Comparisons with State-of-the-art Methods}
\vspace{-0.15cm}

\paragraph{Performance Comparison with Generalist Models.} 
As to multimodal large models, LLaVA-v1.5~\citep{llava/liu2024visual}, mPLUG-Owl2~\citep{mplug-owl2/ye2023mplug}, Qwen-VL~\citep{qwen-vl/bai2023qwen}, and GPT-4V~\citep{openai2023gpt4v} exhibit strong cross-modal reasoning abilities for general tasks. However, when applied to solve geometry tasks, these models are insufficient. 
Our GeoX significantly outperforms these generalists on various geometry datasets, including GeoQA~\citep{geoqa_ngs/chen2021geoqa}, UniGeo~\citep{unigeo_geoformer/chen2022unigeo}, Geometry3K~\citep{inter-gps/lu2021inter}, and PGPS9K~\citep{pgpsnet/zhang2023multi}. 
As indicated in \cref{exp:geoqa} and \cref{table:unigeo}, GeoX achieves top-1 accuracies of 54.9\% and 54.4\%, respectively, significantly outperforming the best generalist models.
%
%
Similarly, on Geometry3K and PGPS9K in~\cref{table:3k9k}, GeoX achieves 58.6\% and 52.7\% in Completion, respectively. In comparison, GPT-4V~\citep{openai2023gpt4v} achieves 34.8\% and 33.3\%, while other models such as Qwen-VL~\citep{qwen-vl/bai2023qwen} and LLaVA~\citep{llava/liu2024visual} perform worse.

\paragraph{Performance Comparison with Specialist Models.}
Compared with geometry specialists such as NGS~\citep{geoqa_ngs/chen2021geoqa}, UniMath-T5~\citep{unimath/liang2023unimath}, Geoformer~\citep{unigeo_geoformer/chen2022unigeo}, DPE-NGS~\citep{dpe-gps/cao2022augmented}, and SCA-GPS~\citep{sca-gps/ning2023symbolic}, GeoX demonstrates superior performance across GeoQA and UniGeo. 
Specifically, GeoX surpasses the best geometry specialist by +4.9\% and +7.6\% on GeoQA and UniGeo-Calculation, respectively. Additionally, our model achieves significant improvements over previous methods on UniGeo-proving by +14.8\% and +43.1\% in \cref{table:unigeo}.
%
%
As reported in \cref{table:3k9k}, our method outperforms SOTA models on Geometry3K and PGPS9K. Notably, previous works~\citep{pgpsnet/zhang2023multi, lans/zhang2023lans} require additional image annotations (Diagram GT) as input, which is labor-consuming and contrary to experimental settings.
To make a fair comparison, we remove Diagram GT and replicate these methods under the original conditions.
%
%
Particularly, we fine-tune LLaVA~\citep{llava/liu2024visual} with formal language and adopt solvers for problem-solving, consistent with the approach used in GeoX.
Extensive results in \cref{exp:geoqa,table:unigeo,table:3k9k} demonstrate the effectiveness of GeoX, achieving state-of-the-art performance across diverse scenarios.

Besides, it should be noted that G-LLaVA-7B~\citep{g-llava/gao2023g} and MAVIS~\citep{mavis/zhang2024mavis} achieve 64.2\% and 66.7\% accuracy on GeoQA. However, these models can produce correct results despite errors in the solving process. In contrast, our method treats any process errors as incorrect results. To this end, we introduce a comparable metric, with detailed results provided in \cref{app:sec:study}. 

\subsection{Quantitative Evaluation on the GPS Task of MathVista}
\vspace{-15pt}
\begin{minipage}[b]{0.6\linewidth}
We provide a quantitative comparison with the model that performed best on the GPS task in MathVista~\citep{mathvista/lumathvista}. To this end, we extract the Geometry subset from MathVista, referred to as MathVista-GEO. 
We assess these methods using the same evaluation script as MathVista, along with the evaluation strategy introduced in \cref{app:sec:study}.
As reported in \cref{table:mathvista}, GeoX is more effective in solving geometry tasks.
\end{minipage}
\hfill
\begin{minipage}[b]{0.38\linewidth}
    \centering
    \begin{table}[H] 
        \caption{
            Accuracy scores on \textit{testmini} of MathVista-GEO.
        }
        \centering
        \resizebox{\textwidth}{!}{
            \centering
            \begin{tabular}{cc}
            \toprule
            Methods & Accuracy \\ 
            \hline
            GPT-4V~\citep{openai2023gpt4v} & 54.8 \\
            GPT-4o~\citep{openai2024gpt4o} & 66.1\\
            \rowcolor[gray]{0.9} 
            GeoX (Ours) &  \textbf{72.6} \\

            \bottomrule
            \end{tabular}
        }
        \label{table:mathvista}
    \end{table}
\end{minipage}

\begin{figure}[h]
\vspace{-5pt}
    \centering
    \includegraphics[width=0.8\linewidth]{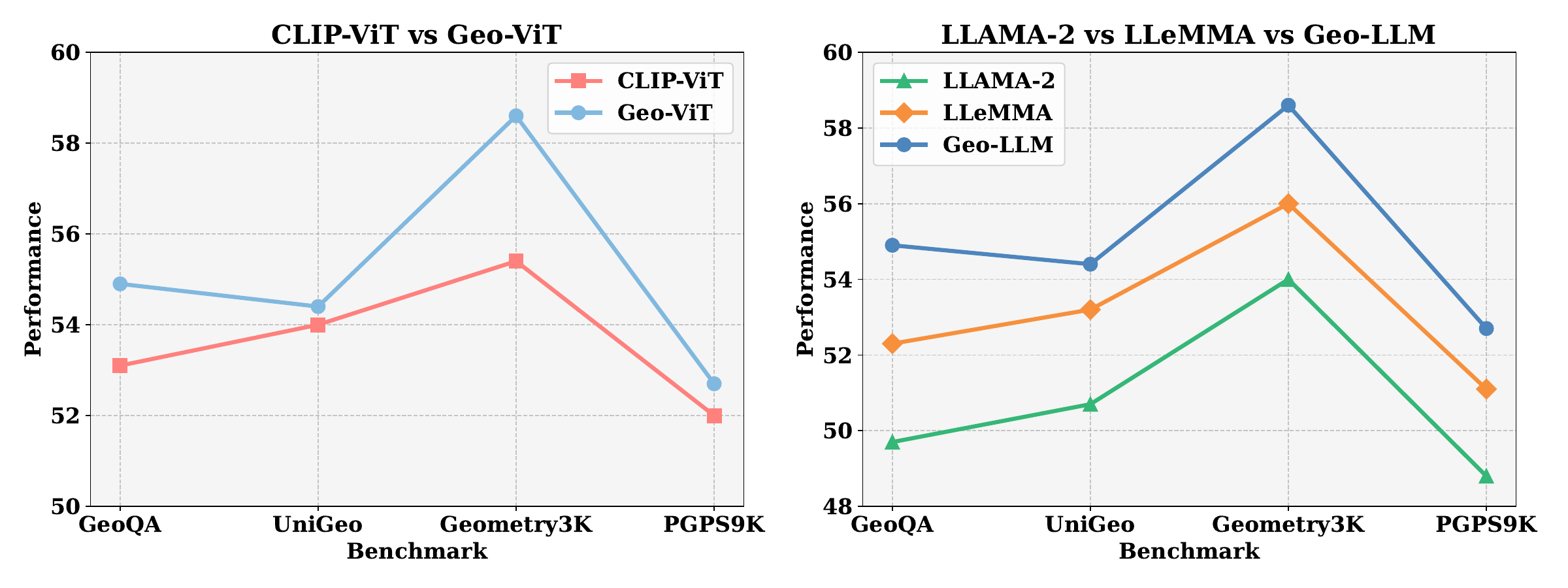}
    \vspace{-12pt}
    \caption{Effectiveness of Uni-modal Pre-training. We compare the widely used CLIP-ViT-B and our Geo-ViT-B, along with three LLM models: LLAMA-2-7B, LLEMMA-7B, and our Geo-LLM-7B.
    }
    \label{fig:abl_pretrain}
    \vspace{-5pt}
\end{figure}

\subsection{Insightful Analyses}
\vspace{-0.15cm}

\paragraph{Effectiveness of Uni-modal Pre-training.} 
We compare Geo-ViT with CLIP-ViT~\citep{clip/radford2021learning}, which has been widely used for GPS in previous studies~\citep{g-llava/gao2023g}. Additionally, we evaluate the performance of different language models in solving geometric problems, including LLAMA-2-7B, LLEMMA-7B, and our Geo-LLM-7B.
As reported in~\cref{fig:abl_pretrain}, compared to general-purpose models or the mathematical model, our pre-trained model demonstrates superior results across various geometry benchmarks.

\paragraph{Effectiveness of Geometry-Language Alignment.}
As illustrated in \cref{table:alignment}, without multi-modal feature alignment, the baseline model perform poorly, achieving only 33.1\% Completion on Geometry3K. The introduction of GS-Former significantly boosts performance. Moreover, our results reveal that formal language is more effective for GPS than natural language, with +2.9\% improvement in Completion on Geometry3K.

\begin{figure}[!t]
    \centering
    \includegraphics[width=0.85\linewidth]{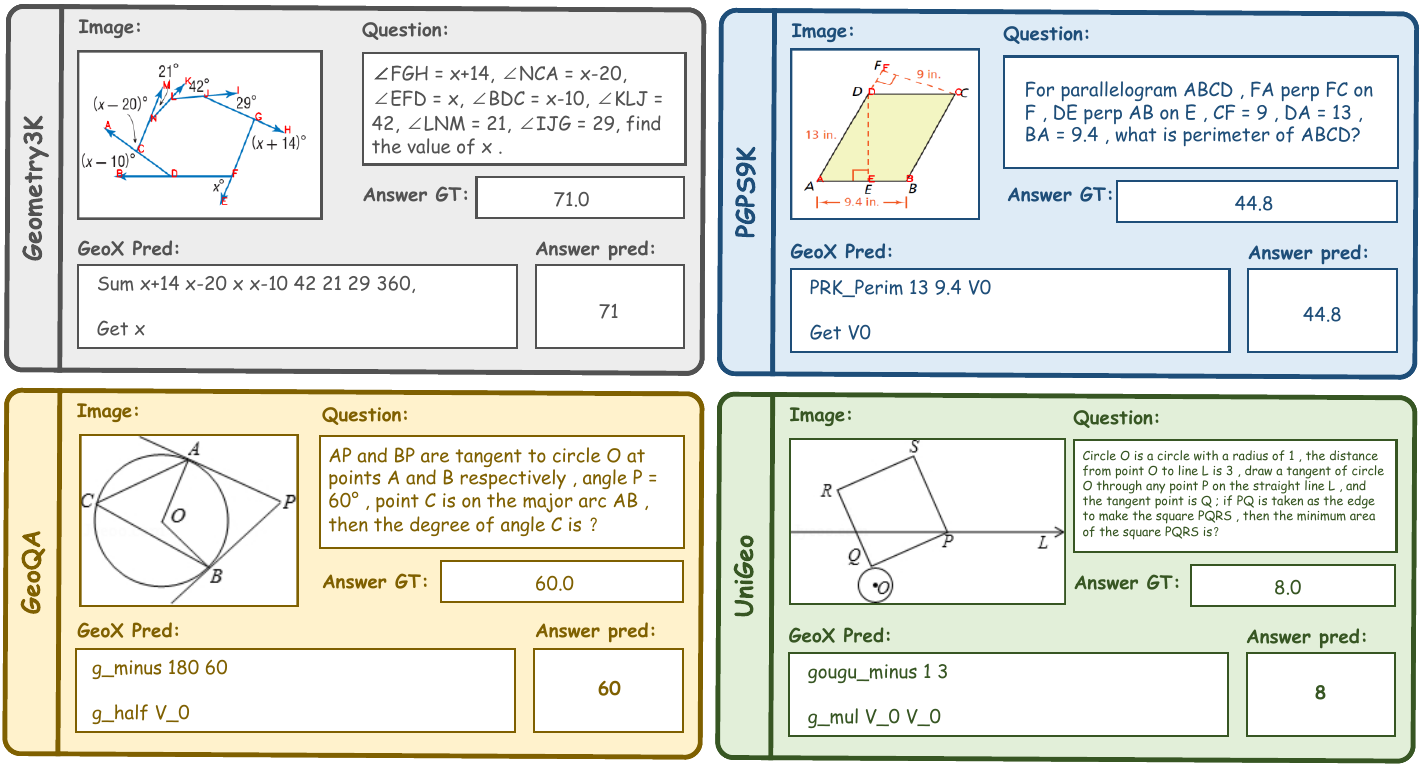}
    \vspace{-6pt}
    \caption{Visualization results on four datasets by our GeoX.
    }
    \label{fig:vis_result}
\vspace{-6pt}
\end{figure}

\begin{table*}[h]
\vspace{-5pt}
    \caption{Effectiveness of geometry-language alignment.}
    \centering
    \begin{minipage}{0.97\textwidth}
        \flushright
        \makeatletter\def\@captype{table}
        \resizebox{\textwidth}{!}{
        \begin{tabular}{lccccccccc}
        \toprule
        \multirow{2}{*}{Module} &\multirow{2}{*}{Alignment} &\multirow{2}{*}{Language} & \multicolumn{3}{c}{Geometry3K} && \multicolumn{3}{c}{PGPS9K} \\
        \cline{4-6} \cline{8-10}
        &&& $Completion\uparrow$ & $Choice\uparrow$ & $Top-3\uparrow$ && $Completion\uparrow$ & $Choice\uparrow$ & $Top-3\uparrow$ \\
        \midrule
        - &$\times$ &- &33.1 & 54.0 & 48.2 &&31.5  & 43.6 &50.1  \\
         \cline{1-1}
          \multirow{3}{*}{GS-Former} & $\times$ &-& 48.6 & 65.7& 63.2 &  &42.7 & 54.3 & 56.8   \\
          & $\checkmark$ &Natural& 55.7 & 71.5 &  67.2 &  & 52.2 & 62.2  & \textbf{67.1}    \\
          &$\checkmark$ &Formal& \textbf{58.6} & \textbf{72.5} & \textbf{69.4}   && \textbf{52.7} & \textbf{63.3} & 65.4 \\
        \bottomrule
        \end{tabular}
        }
        \label{table:alignment}
    \end{minipage}
\vspace{-5pt}
\end{table*}

\begin{table*}[h]
    \caption{Ablation study of modules in GS-Former, assessing the contribution of GQG and SGS modules when GS-Former is utilized for geometry-formal language alignment.}
    \centering
    \begin{minipage}{0.97\textwidth}
        \flushright
        \makeatletter\def\@captype{table}
        \resizebox{\textwidth}{!}{
        \begin{tabular}{ccccccccc}
        \toprule
        \multirow{2}{*}{\begin{tabular}[c]{@{}c@{}}Geo-aware \\ Query Generator\end{tabular}} & \multirow{2}{*}{\begin{tabular}[c]{@{}c@{}}Semantics-guided \\ Geometry Sampler\end{tabular}} & \multicolumn{3}{c}{Geometry3K} && \multicolumn{3}{c}{PGPS9K} \\
        \cline{3-5} \cline{7-9}
        && $Completion\uparrow$ & $Choice\uparrow$ & $Top-3\uparrow$ && $Completion\uparrow$ & $Choice\uparrow$ & $Top-3\uparrow$ \\
        \midrule
         $\times$ &$\times$ & 55.0  & 70.3  &  68.3 &&   49.8 &  59.9 &  64.6  \\ 
        $\checkmark$ &$\times$ & 57.4  & 71.7  & 68.1  && 50.8  & 62.0  &  64.3  \\ 
        $\checkmark$ &$\checkmark$ &\textbf{58.6} & \textbf{72.5} & \textbf{69.4}   && \textbf{52.7} & \textbf{63.3} & \textbf{65.4}   \\ 
        \bottomrule
        \end{tabular}
        }
        \label{table:geoformer}
    \end{minipage}
\vspace{-2pt}
\end{table*}

\paragraph{Ablation of Modules in GS-Former.}
The results in \cref{table:geoformer} demonstrate the effectiveness of the Geo-aware Query Generator (GQG) and Semantics-guided Geometry Sampler (SGS) within GS-Former. Adding the GQG improves Completion by +2.4\% and +1.0\%, while combining both designs yields the best performance. The quantitative results in \cref{app:sec:visualization} further demonstrate GS-Former's effectiveness in capturing valuable information from geometry diagrams, such as lines and symbols.

\vspace{-2pt}
\subsection{Case Study}
\vspace{-2pt}

As shown in \cref{fig:vis_result}, we conduct a case study to analyze the capabilities of GeoX. GeoX tries to predict formalized program sequences composed of mathematical variables, constants, and operations, such as summation (\texttt{Sum}), subtraction (\texttt{g\_minus}), perimeter calculation (\texttt{PRK\_Perim}), the Pythagorean theorem (\texttt{gougu\_minus}), \textit{etc.}, which can be compiled and solved by the GPS-solver.

\begin{figure}[!tb]
    \vspace{-8pt}
    \centering
    \includegraphics[width=0.95\linewidth]{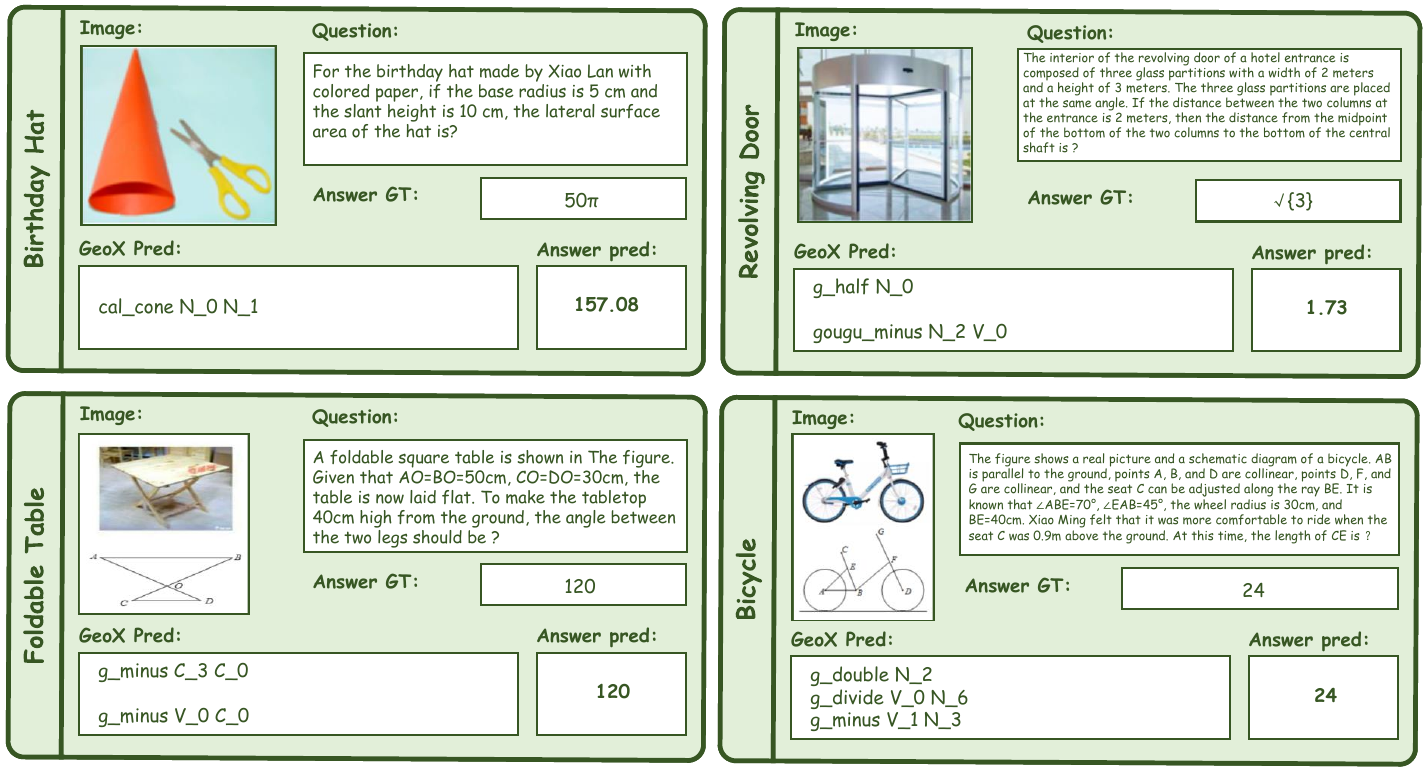}
    \vspace{-6pt}
    \caption{Four visualized examples of geometric problem in natural images solved by our GeoX.
    }
    \label{fig:natural_image}
\end{figure}

Furthermore, we have conducted the generalization validation of GeoX in a broader scope, including its application to geometric problem-solving from natural images. Our GeoX has demonstrated promising performance in these scenarios, indicating the potential for its generalization to even wider fields. We present some visualized examples in \cref{fig:natural_image}.

\section{Conclusion, Limitations, and Future Work}

In this paper, we have proposed GeoX, a novel multi-modal large model specifically designed for automatic Geometry Problem Solving (GPS) tasks. GeoX verifies that formalized vision-language learning is beneficial to learn informative representations for automatic GPS tasks. GeoX can produce formalized process descriptions, which enhance the interpretability of GPS and the correctness of the solution process. Besides, extensive experimental analyses demonstrate GeoX's general capabilities on multiple geometric datasets.





\bibliography{reference}

\begin{thebibliography}{65}
\providecommand{\natexlab}[1]{#1}
\providecommand{\url}[1]{\texttt{#1}}
\expandafter\ifx\csname urlstyle\endcsname\relax
  \providecommand{\doi}[1]{doi: #1}\else
  \providecommand{\doi}{doi: \begingroup \urlstyle{rm}\Url}\fi

\bibitem[Achiam et~al.(2023)Achiam, Adler, Agarwal, Ahmad, Akkaya, Aleman, Almeida, Altenschmidt, Altman, Anadkat, et~al.]{gpt-4v/achiam2023gpt}
Josh Achiam, Steven Adler, Sandhini Agarwal, Lama Ahmad, Ilge Akkaya, Florencia~Leoni Aleman, Diogo Almeida, Janko Altenschmidt, Sam Altman, Shyamal Anadkat, et~al.
\newblock Gpt-4 technical report.
\newblock \emph{arXiv preprint arXiv:2303.08774}, 2023.

\bibitem[Alayrac et~al.(2022)Alayrac, Donahue, Luc, Miech, Barr, Hasson, Lenc, Mensch, Millican, Reynolds, et~al.]{flamingo/alayrac2022flamingo}
Jean-Baptiste Alayrac, Jeff Donahue, Pauline Luc, Antoine Miech, Iain Barr, Yana Hasson, Karel Lenc, Arthur Mensch, Katherine Millican, Malcolm Reynolds, et~al.
\newblock Flamingo: a visual language model for few-shot learning.
\newblock \emph{Advances in neural information processing systems}, 35:\penalty0 23716--23736, 2022.

\bibitem[Anthropic(2024)]{2024claude}
Anthropic.
\newblock The claude 3 model family: Opus, sonnet, haiku.
\newblock \url{https://www.anthropic.com,}, 2024.

\bibitem[Azerbayev et~al.(2023)Azerbayev, Schoelkopf, Paster, Dos~Santos, McAleer, Jiang, Deng, Biderman, and Welleck]{llemma/azerbayev2023llemma}
Zhangir Azerbayev, Hailey Schoelkopf, Keiran Paster, Marco Dos~Santos, Stephen~Marcus McAleer, Albert~Q Jiang, Jia Deng, Stella Biderman, and Sean Welleck.
\newblock Llemma: An open language model for mathematics.
\newblock In \emph{The Twelfth International Conference on Learning Representations}, 2023.

\bibitem[Bai et~al.(2023)Bai, Bai, Yang, Wang, Tan, Wang, Lin, Zhou, and Zhou]{qwen-vl/bai2023qwen}
Jinze Bai, Shuai Bai, Shusheng Yang, Shijie Wang, Sinan Tan, Peng Wang, Junyang Lin, Chang Zhou, and Jingren Zhou.
\newblock Qwen-vl: A frontier large vision-language model with versatile abilities.
\newblock \emph{arXiv preprint arXiv:2308.12966}, 2023.

\bibitem[Bianco et~al.(2023)Bianco, Celona, Donzella, and Napoletano]{llm_caption1/bianco2023improving}
Simone Bianco, Luigi Celona, Marco Donzella, and Paolo Napoletano.
\newblock Improving image captioning descriptiveness by ranking and llm-based fusion.
\newblock \emph{arXiv preprint arXiv:2306.11593}, 2023.

\bibitem[Brown(2020)]{gpt3/brown2020language}
Tom~B Brown.
\newblock Language models are few-shot learners.
\newblock \emph{arXiv preprint ArXiv:2005.14165}, 2020.

\bibitem[Caffagni et~al.(2024)Caffagni, Cocchi, Moratelli, Sarto, Cornia, Baraldi, and Cucchiara]{wiki-llava/caffagni2024wiki}
Davide Caffagni, Federico Cocchi, Nicholas Moratelli, Sara Sarto, Marcella Cornia, Lorenzo Baraldi, and Rita Cucchiara.
\newblock Wiki-llava: Hierarchical retrieval-augmented generation for multimodal llms.
\newblock In \emph{Proceedings of the IEEE/CVF Conference on Computer Vision and Pattern Recognition}, pp.\  1818--1826, 2024.

\bibitem[Cao \& Xiao(2022)Cao and Xiao]{dpe-gps/cao2022augmented}
Jie Cao and Jing Xiao.
\newblock An augmented benchmark dataset for geometric question answering through dual parallel text encoding.
\newblock In \emph{Proceedings of the 29th International Conference on Computational Linguistics}, pp.\  1511--1520, 2022.

\bibitem[Chen et~al.(2021)Chen, Tang, Qin, Liang, Liu, Xing, and Lin]{geoqa_ngs/chen2021geoqa}
Jiaqi Chen, Jianheng Tang, Jinghui Qin, Xiaodan Liang, Lingbo Liu, Eric Xing, and Liang Lin.
\newblock Geoqa: A geometric question answering benchmark towards multimodal numerical reasoning.
\newblock In \emph{Findings of the Association for Computational Linguistics: ACL-IJCNLP 2021}, pp.\  513--523, 2021.

\bibitem[Chen et~al.(2022)Chen, Li, Qin, Lu, Lin, Chen, and Liang]{unigeo_geoformer/chen2022unigeo}
Jiaqi Chen, Tong Li, Jinghui Qin, Pan Lu, Liang Lin, Chongyu Chen, and Xiaodan Liang.
\newblock Unigeo: Unifying geometry logical reasoning via reformulating mathematical expression.
\newblock In \emph{Proceedings of the 2022 Conference on Empirical Methods in Natural Language Processing}, pp.\  3313--3323, 2022.

\bibitem[Chen et~al.(2024{\natexlab{a}})Chen, Chen, Zhang, Li, Yu, Fei, Zhu, Fan, and Chen]{ll3da/chen2024ll3da}
Sijin Chen, Xin Chen, Chi Zhang, Mingsheng Li, Gang Yu, Hao Fei, Hongyuan Zhu, Jiayuan Fan, and Tao Chen.
\newblock Ll3da: Visual interactive instruction tuning for omni-3d understanding reasoning and planning.
\newblock In \emph{Proceedings of the IEEE/CVF Conference on Computer Vision and Pattern Recognition}, pp.\  26428--26438, 2024{\natexlab{a}}.

\bibitem[Chen et~al.(2024{\natexlab{b}})Chen, Wang, Tian, Ye, Gao, Cui, Tong, Hu, Luo, Ma, et~al.]{chen2024far}
Zhe Chen, Weiyun Wang, Hao Tian, Shenglong Ye, Zhangwei Gao, Erfei Cui, Wenwen Tong, Kongzhi Hu, Jiapeng Luo, Zheng Ma, et~al.
\newblock How far are we to gpt-4v? closing the gap to commercial multimodal models with open-source suites.
\newblock \emph{arXiv preprint arXiv:2404.16821}, 2024{\natexlab{b}}.

\bibitem[Dai et~al.(2023)Dai, Li, Li, Tiong, Zhao, Wang, Li, Fung, and Hoi]{instructblip/instructblip}
Wenliang Dai, Junnan Li, Dongxu Li, Anthony Meng~Huat Tiong, Junqi Zhao, Weisheng Wang, Boyang Li, Pascale Fung, and Steven Hoi.
\newblock Instructblip: Towards general-purpose vision-language models with instruction tuning, 2023.

\bibitem[Faulstich \& Oster(2024)Faulstich and Oster]{algebraic/faulstich2024coupled}
Fabian~M Faulstich and Mathias Oster.
\newblock Coupled cluster theory: Toward an algebraic geometry formulation.
\newblock \emph{SIAM Journal on Applied Algebra and Geometry}, 8\penalty0 (1):\penalty0 138--188, 2024.

\bibitem[Fukunaga(2013)]{statistical}
Keinosuke Fukunaga.
\newblock \emph{Introduction to statistical pattern recognition}.
\newblock Elsevier, 2013.

\bibitem[Gao et~al.(2023)Gao, Pi, Zhang, Ye, Zhong, Wang, Hong, Han, Xu, Li, et~al.]{g-llava/gao2023g}
Jiahui Gao, Renjie Pi, Jipeng Zhang, Jiacheng Ye, Wanjun Zhong, Yufei Wang, Lanqing Hong, Jianhua Han, Hang Xu, Zhenguo Li, et~al.
\newblock G-llava: Solving geometric problem with multi-modal large language model.
\newblock \emph{arXiv preprint arXiv:2312.11370}, 2023.

\bibitem[Han et~al.(2024)Han, Gong, Zhang, Wang, Zhang, Lin, Qiao, Gao, and Yue]{one-llm/han2024onellm}
Jiaming Han, Kaixiong Gong, Yiyuan Zhang, Jiaqi Wang, Kaipeng Zhang, Dahua Lin, Yu~Qiao, Peng Gao, and Xiangyu Yue.
\newblock Onellm: One framework to align all modalities with language.
\newblock In \emph{Proceedings of the IEEE/CVF Conference on Computer Vision and Pattern Recognition}, pp.\  26584--26595, 2024.

\bibitem[Hao et~al.(2022)Hao, Zhang, Yin, and Huang]{pgdp5k/hao2022pgdp5k}
Yihan Hao, Mingliang Zhang, Fei Yin, and Lin-Lin Huang.
\newblock Pgdp5k: A diagram parsing dataset for plane geometry problems.
\newblock In \emph{2022 26th International Conference on Pattern Recognition (ICPR)}, pp.\  1763--1769. IEEE, 2022.

\bibitem[He et~al.(2022)He, Chen, Xie, Li, Doll{\'a}r, and Girshick]{mae/he2022masked}
Kaiming He, Xinlei Chen, Saining Xie, Yanghao Li, Piotr Doll{\'a}r, and Ross Girshick.
\newblock Masked autoencoders are scalable vision learners.
\newblock In \emph{Proceedings of the IEEE/CVF conference on computer vision and pattern recognition}, pp.\  16000--16009, 2022.

\bibitem[Imani et~al.(2023)Imani, Du, and Shrivastava]{mathprompter/imani2023mathprompter}
Shima Imani, Liang Du, and Harsh Shrivastava.
\newblock Mathprompter: Mathematical reasoning using large language models.
\newblock \emph{arXiv preprint arXiv:2303.05398}, 2023.

\bibitem[Jang et~al.(2016)Jang, Gu, and Poole]{gumbel-softmax/jang2016categorical}
Eric Jang, Shixiang Gu, and Ben Poole.
\newblock Categorical reparameterization with gumbel-softmax.
\newblock \emph{arXiv preprint arXiv:1611.01144}, 2016.

\bibitem[Kenton \& Toutanova(2019)Kenton and Toutanova]{bert/kenton2019bert}
Jacob Devlin Ming-Wei~Chang Kenton and Lee~Kristina Toutanova.
\newblock Bert: Pre-training of deep bidirectional transformers for language understanding.
\newblock In \emph{Proceedings of NAACL-HLT}, pp.\  4171--4186, 2019.

\bibitem[Li et~al.(2023{\natexlab{a}})Li, Zhang, Chen, Wang, Pu, Yang, Li, and Liu]{mimic/li2023mimic}
Bo~Li, Yuanhan Zhang, Liangyu Chen, Jinghao Wang, Fanyi Pu, Jingkang Yang, Chunyuan Li, and Ziwei Liu.
\newblock Mimic-it: Multi-modal in-context instruction tuning.
\newblock \emph{arXiv preprint arXiv:2306.05425}, 2023{\natexlab{a}}.

\bibitem[Li et~al.(2023{\natexlab{b}})Li, Li, Savarese, and Hoi]{blip-2/li2023blip}
Junnan Li, Dongxu Li, Silvio Savarese, and Steven Hoi.
\newblock Blip-2: Bootstrapping language-image pre-training with frozen image encoders and large language models.
\newblock In \emph{International conference on machine learning}, pp.\  19730--19742. PMLR, 2023{\natexlab{b}}.

\bibitem[Li et~al.(2023{\natexlab{c}})Li, Chen, Zhang, Chen, Zhu, Yin, Yu, and Chen]{m3dbench/li2023m3dbench}
Mingsheng Li, Xin Chen, Chi Zhang, Sijin Chen, Hongyuan Zhu, Fukun Yin, Gang Yu, and Tao Chen.
\newblock M3dbench: Let's instruct large models with multi-modal 3d prompts.
\newblock \emph{arXiv preprint arXiv:2312.10763}, 2023{\natexlab{c}}.

\bibitem[Li et~al.(2024)Li, Du, Liu, Zhang, Liu, Zhang, and Cai]{eagle/li2024eagle}
Zhihao Li, Yao Du, Yang Liu, Yan Zhang, Yufang Liu, Mengdi Zhang, and Xunliang Cai.
\newblock Eagle: Elevating geometric reasoning through llm-empowered visual instruction tuning.
\newblock \emph{arXiv preprint arXiv:2408.11397}, 2024.

\bibitem[Liang et~al.(2023)Liang, Yang, Zhang, and Zhang]{unimath/liang2023unimath}
Zhenwen Liang, Tianyu Yang, Jipeng Zhang, and Xiangliang Zhang.
\newblock Unimath: A foundational and multimodal mathematical reasoner.
\newblock In \emph{Proceedings of the 2023 Conference on Empirical Methods in Natural Language Processing}, pp.\  7126--7133, 2023.

\bibitem[Liu et~al.(2024)Liu, Li, Wu, and Lee]{llava/liu2024visual}
Haotian Liu, Chunyuan Li, Qingyang Wu, and Yong~Jae Lee.
\newblock Visual instruction tuning.
\newblock \emph{Advances in neural information processing systems}, 36, 2024.

\bibitem[Lu et~al.()Lu, Bansal, Xia, Liu, Li, Hajishirzi, Cheng, Chang, Galley, and Gao]{mathvista/lumathvista}
Pan Lu, Hritik Bansal, Tony Xia, Jiacheng Liu, Chunyuan Li, Hannaneh Hajishirzi, Hao Cheng, Kai-Wei Chang, Michel Galley, and Jianfeng Gao.
\newblock Mathvista: Evaluating mathematical reasoning of foundation models in visual contexts.
\newblock In \emph{The Twelfth International Conference on Learning Representations}.

\bibitem[Lu et~al.(2021)Lu, Gong, Jiang, Qiu, Huang, Liang, and Zhu]{inter-gps/lu2021inter}
Pan Lu, Ran Gong, Shibiao Jiang, Liang Qiu, Siyuan Huang, Xiaodan Liang, and Song-Chun Zhu.
\newblock Inter-gps: Interpretable geometry problem solving with formal language and symbolic reasoning.
\newblock \emph{arXiv preprint arXiv:2105.04165}, 2021.

\bibitem[Ning et~al.(2023)Ning, Wang, Huang, and Huang]{sca-gps/ning2023symbolic}
Maizhen Ning, Qiu-Feng Wang, Kaizhu Huang, and Xiaowei Huang.
\newblock A symbolic characters aware model for solving geometry problems.
\newblock In \emph{Proceedings of the 31st ACM International Conference on Multimedia}, pp.\  7767--7775, 2023.

\bibitem[OpenAI(2023)]{openai2023gpt4v}
OpenAI.
\newblock Gpt-4v.
\newblock \url{https://openai.com/index/gpt-4v-system-card/}, 2023.

\bibitem[OpenAI(2024)]{openai2024gpt4o}
OpenAI.
\newblock Hello gpt-4o.
\newblock \url{https://openai.com/index/hello-gpt-4o/}, 2024.

\bibitem[Ouyang et~al.(2022)Ouyang, Wu, Jiang, Almeida, Wainwright, Mishkin, Zhang, Agarwal, Slama, Ray, et~al.]{chatgpt/ouyang2022training}
Long Ouyang, Jeffrey Wu, Xu~Jiang, Diogo Almeida, Carroll Wainwright, Pamela Mishkin, Chong Zhang, Sandhini Agarwal, Katarina Slama, Alex Ray, et~al.
\newblock Training language models to follow instructions with human feedback.
\newblock \emph{Advances in neural information processing systems}, 35:\penalty0 27730--27744, 2022.

\bibitem[Peng et~al.(2023)Peng, Fu, Liang, Gao, and Tang]{geodrl/peng2023geodrl}
Shuai Peng, Di~Fu, Yijun Liang, Liangcai Gao, and Zhi Tang.
\newblock Geodrl: A self-learning framework for geometry problem solving using reinforcement learning in deductive reasoning.
\newblock In \emph{Findings of the Association for Computational Linguistics: ACL 2023}, pp.\  13468--13480, 2023.

\bibitem[Peng et~al.(2024)Peng, Li, Zhou, Xia, Zhang, Bai, Mao, Wang, He, Zhou, et~al.]{peng2024chimera}
Tianshuo Peng, Mingsheng Li, Hongbin Zhou, Renqiu Xia, Renrui Zhang, Lei Bai, Song Mao, Bin Wang, Conghui He, Aojun Zhou, et~al.
\newblock Chimera: Improving generalist model with domain-specific experts.
\newblock \emph{arXiv preprint arXiv:2412.05983}, 2024.

\bibitem[Perez et~al.(2018)Perez, Strub, De~Vries, Dumoulin, and Courville]{film/perez2018film}
Ethan Perez, Florian Strub, Harm De~Vries, Vincent Dumoulin, and Aaron Courville.
\newblock Film: Visual reasoning with a general conditioning layer.
\newblock In \emph{Proceedings of the AAAI conference on artificial intelligence}, volume~32, 2018.

\bibitem[Radford et~al.(2021)Radford, Kim, Hallacy, Ramesh, Goh, Agarwal, Sastry, Askell, Mishkin, Clark, et~al.]{clip/radford2021learning}
Alec Radford, Jong~Wook Kim, Chris Hallacy, Aditya Ramesh, Gabriel Goh, Sandhini Agarwal, Girish Sastry, Amanda Askell, Pamela Mishkin, Jack Clark, et~al.
\newblock Learning transferable visual models from natural language supervision.
\newblock In \emph{International conference on machine learning}, pp.\  8748--8763. PMLR, 2021.

\bibitem[Reid et~al.(2024)Reid, Savinov, Teplyashin, Lepikhin, Lillicrap, Alayrac, Soricut, Lazaridou, Firat, Schrittwieser, et~al.]{reid2024gemini}
Machel Reid, Nikolay Savinov, Denis Teplyashin, Dmitry Lepikhin, Timothy Lillicrap, Jean-baptiste Alayrac, Radu Soricut, Angeliki Lazaridou, Orhan Firat, Julian Schrittwieser, et~al.
\newblock Gemini 1.5: Unlocking multimodal understanding across millions of tokens of context.
\newblock \emph{arXiv preprint arXiv:2403.05530}, 2024.

\bibitem[Rotstein et~al.(2023)Rotstein, Bensaid, Brody, Ganz, and Kimmel]{llm_caption2/rotstein2023fusecap}
Noam Rotstein, David Bensaid, Shaked Brody, Roy Ganz, and Ron Kimmel.
\newblock Fusecap: Leveraging large language models to fuse visual data into enriched image captions.
\newblock \emph{arXiv preprint arXiv:2305.17718}, 2023.

\bibitem[Sachan \& Xing(2017)Sachan and Xing]{rule1/sachan2017learning}
Mrinmaya Sachan and Eric Xing.
\newblock Learning to solve geometry problems from natural language demonstrations in textbooks.
\newblock In \emph{Proceedings of the 6th Joint Conference on Lexical and Computational Semantics (* SEM 2017)}, pp.\  251--261, 2017.

\bibitem[Santoro et~al.(2017)Santoro, Raposo, Barrett, Malinowski, Pascanu, Battaglia, and Lillicrap]{rn/santoro2017simple}
Adam Santoro, David Raposo, David~G Barrett, Mateusz Malinowski, Razvan Pascanu, Peter Battaglia, and Timothy Lillicrap.
\newblock A simple neural network module for relational reasoning.
\newblock \emph{Advances in neural information processing systems}, 30, 2017.

\bibitem[Seo et~al.(2015)Seo, Hajishirzi, Farhadi, Etzioni, and Malcolm]{geos/seo2015solving}
Minjoon Seo, Hannaneh Hajishirzi, Ali Farhadi, Oren Etzioni, and Clint Malcolm.
\newblock Solving geometry problems: Combining text and diagram interpretation.
\newblock In \emph{Proceedings of the 2015 conference on empirical methods in natural language processing}, pp.\  1466--1476, 2015.

\bibitem[Shi et~al.(2024)Shi, Hu, Bin, Liu, Yang, Ng, Bing, and Lee]{math-llava/shi2024math}
Wenhao Shi, Zhiqiang Hu, Yi~Bin, Junhua Liu, Yang Yang, See-Kiong Ng, Lidong Bing, and Roy Ka-Wei Lee.
\newblock Math-llava: Bootstrapping mathematical reasoning for multimodal large language models.
\newblock \emph{arXiv preprint arXiv:2406.17294}, 2024.

\bibitem[Team et~al.(2023)Team, Anil, Borgeaud, Wu, Alayrac, Yu, Soricut, Schalkwyk, Dai, Hauth, et~al.]{team2023gemini}
Gemini Team, Rohan Anil, Sebastian Borgeaud, Yonghui Wu, Jean-Baptiste Alayrac, Jiahui Yu, Radu Soricut, Johan Schalkwyk, Andrew~M Dai, Anja Hauth, et~al.
\newblock Gemini: a family of highly capable multimodal models.
\newblock \emph{arXiv preprint arXiv:2312.11805}, 2023.

\bibitem[Team(2023)]{internlm/team2023internlm}
InternLM Team.
\newblock Internlm: A multilingual language model with progressively enhanced capabilities, 2023.

\bibitem[Touvron et~al.(2023{\natexlab{a}})Touvron, Lavril, Izacard, Martinet, Lachaux, Lacroix, Rozi{\`e}re, Goyal, Hambro, Azhar, et~al.]{llama/touvron2023llama}
Hugo Touvron, Thibaut Lavril, Gautier Izacard, Xavier Martinet, Marie-Anne Lachaux, Timoth{\'e}e Lacroix, Baptiste Rozi{\`e}re, Naman Goyal, Eric Hambro, Faisal Azhar, et~al.
\newblock Llama: Open and efficient foundation language models.
\newblock \emph{arXiv preprint arXiv:2302.13971}, 2023{\natexlab{a}}.

\bibitem[Touvron et~al.(2023{\natexlab{b}})Touvron, Martin, Stone, Albert, Almahairi, Babaei, Bashlykov, Batra, Bhargava, Bhosale, et~al.]{llama2/touvron2023llama}
Hugo Touvron, Louis Martin, Kevin Stone, Peter Albert, Amjad Almahairi, Yasmine Babaei, Nikolay Bashlykov, Soumya Batra, Prajjwal Bhargava, Shruti Bhosale, et~al.
\newblock Llama 2: Open foundation and fine-tuned chat models.
\newblock \emph{arXiv preprint arXiv:2307.09288}, 2023{\natexlab{b}}.

\bibitem[Trinh et~al.(2024)Trinh, Wu, Le, He, and Luong]{alphageometry/trinh2024solving}
Trieu~H Trinh, Yuhuai Wu, Quoc~V Le, He~He, and Thang Luong.
\newblock Solving olympiad geometry without human demonstrations.
\newblock \emph{Nature}, 625\penalty0 (7995):\penalty0 476--482, 2024.

\bibitem[Wang et~al.(2024{\natexlab{a}})Wang, Gu, Xu, Zhang, Shi, and He]{wang2024unimernet}
Bin Wang, Zhuangcheng Gu, Chao Xu, Bo~Zhang, Botian Shi, and Conghui He.
\newblock Unimernet: A universal network for real-world mathematical expression recognition.
\newblock \emph{arXiv preprint arXiv:2404.15254}, 2024{\natexlab{a}}.

\bibitem[Wang et~al.(2024{\natexlab{b}})Wang, Xu, Zhao, Ouyang, Wu, Zhao, Xu, Liu, Qu, Shang, et~al.]{wang2024mineru}
Bin Wang, Chao Xu, Xiaomeng Zhao, Linke Ouyang, Fan Wu, Zhiyuan Zhao, Rui Xu, Kaiwen Liu, Yuan Qu, Fukai Shang, et~al.
\newblock Mineru: An open-source solution for precise document content extraction.
\newblock \emph{arXiv preprint arXiv:2409.18839}, 2024{\natexlab{b}}.

\bibitem[Wang et~al.(2022)Wang, Guo, Deng, and Lu]{rethinking}
Haoqing Wang, Xun Guo, Zhi-Hong Deng, and Yan Lu.
\newblock Rethinking minimal sufficient representation in contrastive learning.
\newblock In \emph{Proceedings of the IEEE/CVF Conference on Computer Vision and Pattern Recognition}, pp.\  16041--16050, 2022.

\bibitem[Wang et~al.()Wang, Wang, Zhou, Li, Hua, Tang, et~al.]{retrieval/wang2024multimodal}
Yabing Wang, Le~Wang, Qiang Zhou, Hao Li, Gang Hua, Wei Tang, et~al.
\newblock Multimodal llm enhanced cross-lingual cross-modal retrieval.
\newblock In \emph{ACM Multimedia 2024}.

\bibitem[Wu \& Xie(2024)Wu and Xie]{vqa1/wu2024v}
Penghao Wu and Saining Xie.
\newblock V?: Guided visual search as a core mechanism in multimodal llms.
\newblock In \emph{Proceedings of the IEEE/CVF Conference on Computer Vision and Pattern Recognition}, pp.\  13084--13094, 2024.

\bibitem[Wu et~al.(2024)Wu, Zhang, Liu, Tang, Wang, Wang, and Wang]{e-gps/wu2024gps}
Wenjun Wu, Lingling Zhang, Jun Liu, Xi~Tang, Yaxian Wang, Shaowei Wang, and Qianying Wang.
\newblock E-gps: Explainable geometry problem solving via top-down solver and bottom-up generator.
\newblock In \emph{Proceedings of the IEEE/CVF Conference on Computer Vision and Pattern Recognition}, pp.\  13828--13837, 2024.

\bibitem[Xia et~al.(2023)Xia, Zhang, Peng, Liao, Ye, Shi, Yan, and Qiao]{structchart/xia2023structchart}
Renqiu Xia, Bo~Zhang, Haoyang Peng, Ning Liao, Peng Ye, Botian Shi, Junchi Yan, and Yu~Qiao.
\newblock Structchart: Perception, structuring, reasoning for visual chart understanding.
\newblock \emph{arXiv preprint arXiv:2309.11268}, 2023.

\bibitem[Xia et~al.(2024{\natexlab{a}})Xia, Mao, Yan, Zhou, Zhang, Peng, Pi, Fu, Wu, Ye, et~al.]{xia2024docgenome}
Renqiu Xia, Song Mao, Xiangchao Yan, Hongbin Zhou, Bo~Zhang, Haoyang Peng, Jiahao Pi, Daocheng Fu, Wenjie Wu, Hancheng Ye, et~al.
\newblock Docgenome: An open large-scale scientific document benchmark for training and testing multi-modal large language models.
\newblock \emph{arXiv preprint arXiv:2406.11633}, 2024{\natexlab{a}}.

\bibitem[Xia et~al.(2024{\natexlab{b}})Xia, Zhang, Ye, Yan, Liu, Zhou, Chen, Dou, Shi, Yan, et~al.]{xia2024chartx}
Renqiu Xia, Bo~Zhang, Hancheng Ye, Xiangchao Yan, Qi~Liu, Hongbin Zhou, Zijun Chen, Min Dou, Botian Shi, Junchi Yan, et~al.
\newblock Chartx \& chartvlm: A versatile benchmark and foundation model for complicated chart reasoning.
\newblock \emph{arXiv preprint arXiv:2402.12185}, 2024{\natexlab{b}}.

\bibitem[Ye et~al.(2023)Ye, Xu, Ye, Yan, Liu, Qian, Zhang, Huang, and Zhou]{mplug-owl2/ye2023mplug}
Qinghao Ye, Haiyang Xu, Jiabo Ye, Ming Yan, Haowei Liu, Qi~Qian, Ji~Zhang, Fei Huang, and Jingren Zhou.
\newblock mplug-owl2: Revolutionizing multi-modal large language model with modality collaboration.
\newblock \emph{arXiv preprint arXiv:2311.04257}, 2023.

\bibitem[Yu et~al.(2019)Yu, Yu, Cui, Tao, and Tian]{macn/yu2019deep}
Zhou Yu, Jun Yu, Yuhao Cui, Dacheng Tao, and Qi~Tian.
\newblock Deep modular co-attention networks for visual question answering.
\newblock In \emph{Proceedings of the IEEE/CVF conference on computer vision and pattern recognition}, pp.\  6281--6290, 2019.

\bibitem[Zhang et~al.(2023{\natexlab{a}})Zhang, Yuan, Shi, Chen, Li, and Qiao]{zhang2023uni3d}
Bo~Zhang, Jiakang Yuan, Botian Shi, Tao Chen, Yikang Li, and Yu~Qiao.
\newblock Uni3d: A unified baseline for multi-dataset 3d object detection.
\newblock In \emph{Proceedings of the IEEE/CVF Conference on Computer Vision and Pattern Recognition}, pp.\  9253--9262, 2023{\natexlab{a}}.

\bibitem[Zhang et~al.(2023{\natexlab{b}})Zhang, Li, Yin, and Liu]{lans/zhang2023lans}
Ming-Liang Zhang, Zhong-Zhi Li, Fei Yin, and Cheng-Lin Liu.
\newblock Lans: A layout-aware neural solver for plane geometry problem.
\newblock \emph{arXiv preprint arXiv:2311.16476}, 2023{\natexlab{b}}.

\bibitem[Zhang et~al.(2023{\natexlab{c}})Zhang, Yin, and Liu]{pgpsnet/zhang2023multi}
Ming-Liang Zhang, Fei Yin, and Cheng-Lin Liu.
\newblock A multi-modal neural geometric solver with textual clauses parsed from diagram.
\newblock \emph{arXiv preprint arXiv:2302.11097}, 2023{\natexlab{c}}.

\bibitem[Zhang et~al.(2024)Zhang, Wei, Jiang, Zhang, Guo, Tong, Liu, Zhou, Wei, Zhang, et~al.]{mavis/zhang2024mavis}
Renrui Zhang, Xinyu Wei, Dongzhi Jiang, Yichi Zhang, Ziyu Guo, Chengzhuo Tong, Jiaming Liu, Aojun Zhou, Bin Wei, Shanghang Zhang, et~al.
\newblock Mavis: Mathematical visual instruction tuning.
\newblock \emph{arXiv preprint arXiv:2407.08739}, 2024.

\end{thebibliography}
\bibliographystyle{iclr2025_conference}


\clearpage
\appendix
\section*{
    {\hfill \LARGE Appendix \hfill }
}
The appendix mainly includes the following aspects:
\begin{itemize}
    \item Sec.~\ref{app:sec:thm}: Theoretical analysis on the proposed formalized vision-language pre-training.
    \item Sec.~\ref{app:sec:visualization}: More visualization results.
    \item Sec.~\ref{app:sec:formal_data}: Examples of formalized diagram-caption pairs.
    \item Sec.~\ref{app:sec:study}: Additional quantitative evaluations.
    \item Sec.~\ref{app:sec:corpus}: Data acquisition for geometric corpus.
    \item Sec.~\ref{app:sec:implementation}: Implementation details. \item Sec.~\ref{app:sec:gpt4v_formalprogram}: Further discussions and analyses.
\end{itemize}

\paragraph{Codes} are released at \url{https://github.com/UniModal4Reasoning/GeoX}, including the process of the training and evaluation of GeoX.

\section{Theoretical Analysis}
\label{sec:proof}

In this section, we theoretically explain why the proposed formalized pre-training benefits more than informal pre-training methods in downstream tasks of geometry problems. First, we consider the sufficient representations for the pre-training of the Geometric Problem-Solving (GPS) models, containing the information shared between different modalities of geometry data. The definition of sufficient representations is borrowed and extended from the idea in ~\citet{rethinking}, We denote $T_{f}$ as the target formalized descriptions of samples in the pre-training dataset, while $T_{inf}$ is denoted as the informal descriptions of samples in the pre-training dataset. The representations learned from $T_{f}$ is denoted as $z_{f}$, while the representations learned from $T_{inf}$ is denoted as $z_{inf}$. The downstream task label is denoted as $T$, which is a formalized textual sequence that will be fed into the GPS-Solver for verifiable numerical solutions.

\noindent\textbf{Definition 1.} \textit{(Sufficient Representations) The representations $z_{1, \text{suf}}$ of $y_1$ is sufficient for another task $y_2$ \textbf{if and only if }$I(z_{1, \text{suf}}, y_2) = I(y_1, y_2)$, where $z_{1, \text{suf}}$ is learned from $y_1$, and $y_1$, $y_2$ are the labels of two different prediction tasks that contain the shared information. $I(\cdot, \cdot)$ refers to the mutual information between the two variables.}

\noindent\textbf{Definition 2.} \textit{(Minimal Sufficient Representations) The representations $z_{1, \text{min}}$ is minimal sufficient \textbf{if and only if }$I(z_{1, \text{min}}, y_2) = \min_{z_{1, \text{suf}}} {I(z_{1, \text{suf}}, y_2)}$.}

\noindent\textbf{Lemma 1.} $z_{f}$ provides more information about the downstream task $T$ than $z_{inf}$. That is, $I(z_{f}, T) \geq I(z_{inf}, T)$.

\textit{Proof.} Since both $T_f$ and $T_{inf}$ are supervised learning tasks, their learned representations $z_f$ and $z_{inf}$ are both sufficient representations. However, since  $T_{inf}$ only contains the semantic information without structural context that is required by the downstream tasks. Therefore, it holds that $I(z_{inf}, T) \leq I(z_{\text{suf}}, T), \forall z_{\text{f}}$ that is sufficient. That is, $z_{inf}$ is a minimal sufficient representation. As for $z_{f}$, it learns information from the formalized description and thus is more related to the downstream tasks. Consequently, we have the relationship between $z_{inf}$ and $z_{f}$ as follows,
\begin{equation}
\begin{aligned}
    I(z_{f}, T) &= I(z_{inf}, T) + [I(T_{f}, T|z_{inf}) - I(T_{f}, T|z_{f})]\\
    &\geq I(z_{inf}, T).
\end{aligned}
\end{equation}

The first equation indicates that the mutual information $I(z_{f}, T)$ can be decomposed into the minimal mutual information $I(z_{inf}, T)$ and the information gap between $I(T_{f}, T|z_{inf})$ and $I(T_{f}, T|z_{f})$, where $I(T_{f}, T|z_{inf})$ refers to the information about $T$ that can be observed from $T_{f}$ on condition of $z_{inf}$. Since $T_{f}$ contains more formalized information related to $T$ and $I(z_{inf}, T_{f}) \leq I(z_{f}, T_{f})$, we can get $I(T_{f}, T|z_{inf}) \geq I(T_{f}, T|z_{f})$. Consequently, $I(z_{f}, T) \geq I(z_{inf}, T)$ holds.

\noindent\textbf{Theorem 1.} The upper bound of error rates in downstream tasks using minimal sufficient representations is higher than that using sufficient representations.

\textit{Proof.} For the downstream tasks, we consider the Bayes error rate \citep{statistical} to estimate the lowest achievable error of the classifier. According to the paper \citep{rethinking}, for arbitrary representations $z$, its Bayes error rate $P_e$ satisfies that, 
\begin{equation}
    P_e \leq 1 - \exp[-H(T)+I(z, T)],
\end{equation}
where $H(T)$ represents the entropy of variable $T$. Since $I(z_{f}, T) \geq I(z_{inf}, T)$, it can be concluded that the upper-bound of $P_{e, f}$ is smaller than that of $P_{e, inf}$. This indicates that ideally $z_{f}$ is expected to achieve better performance than $z_{inf}$ in downstream tasks.

\label{app:sec:thm}

\section{More Visualizations}
\label{app:sec:visualization}

\begin{figure}[htbp]
    \vspace{-4pt}
    \centering
    \includegraphics[width=0.8\linewidth]{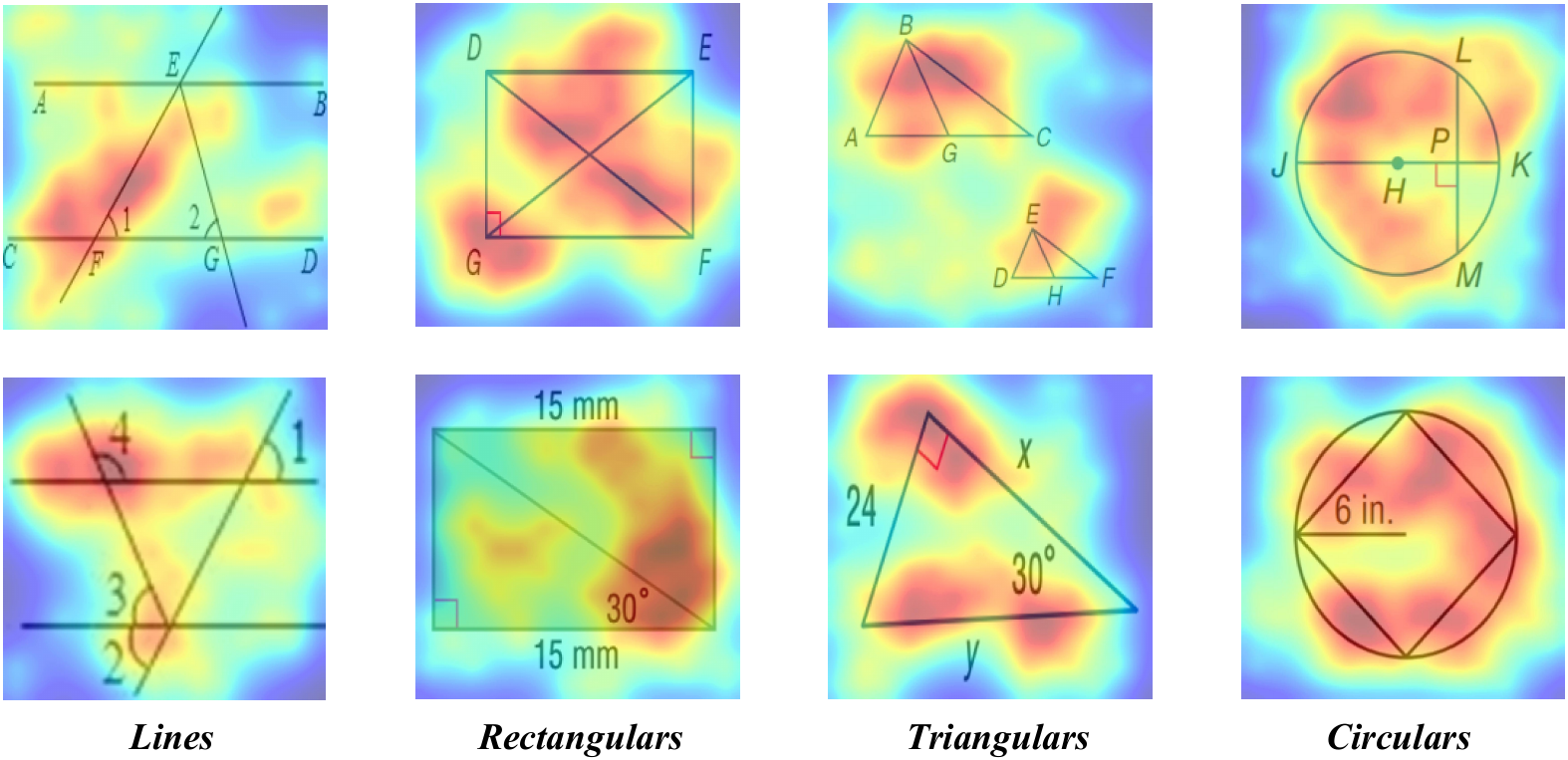}
    \vspace{-6pt}
    \caption{
    Attention map of GS-Former on different types of geometric diagrams.
    }
    \label{fig:attention}
    \vspace{-4pt}
\end{figure}

In \cref{fig:attention}, we present attention maps of GS-Former, which show the model's attention distribution across different regions. The areas with higher brightness indicate regions considered more useful for making decisions. In contrast, darker areas are often semantically irrelevant and uninformative, which will be removed by GS-Former. This visualization highlights our model’s ability to capture pivotal information across complex geometric images, such as lines, rectangles, triangles, circles, etc.

\vspace{-2pt}

\section{Examples of Formalized Diagram-Caption Pairs}
\label{app:sec:formal_data}
\vspace{-2pt}

\begin{table*}[h]
    \centering
    \small
    \resizebox{0.92\textwidth}{!}{  
    \begin{tabular}{c|c|c|c}  
     \toprule
     \textbf{Image} & \textbf{Caption} & \textbf{Image} & \textbf{Caption} \\
     \midrule
     \begin{minipage}[t]{0.2\textwidth}  
     \multirow{8}*{\includegraphics[width=0.98\linewidth]{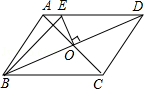}}
     \end{minipage} 
     &
     \begin{minipage}[t]{0.32\textwidth}  
     \texttt{Line A E D}\\
     \texttt{Line A O C}\\
     \texttt{Line B O D}\\
     \texttt{Line B A}\\
     \texttt{Line B C}\\
     \texttt{Line C D}\\
     \texttt{Line B E}\\
     \texttt{Line E O}\\
     \end{minipage} 
     &
     \begin{minipage}[t]{0.2\textwidth}  
     \quad \multirow{8}*{\includegraphics[width=0.74\linewidth]{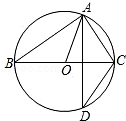}}
     \end{minipage} 
     &
     \begin{minipage}[t]{0.32\textwidth}  
     \texttt{Line B A}\\
     \texttt{Line O A}\\
     \texttt{Line A C}\\
     \texttt{Line B O C}\\
     \texttt{Line A D}\\
     \texttt{Line D C}\\
     \texttt{$\backslash\backslash$odot O lieson A C D B}\\
     \end{minipage} 
     \\
     \midrule

     \begin{minipage}[t]{0.2\textwidth}  
     \quad \multirow{6}*{\includegraphics[width=0.78\linewidth]{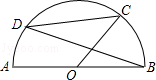}}
     \end{minipage} 
     &
     \begin{minipage}[t]{0.32\textwidth}  
     \texttt{Line A O B}\\
     \texttt{Line D C}\\
     \texttt{Line D B}\\
     \texttt{Line O C}\\
     \texttt{$\backslash\backslash$odot O lieson A D C B}\\
     \end{minipage} 
     &
     \begin{minipage}[t]{0.2\textwidth}  
     \quad \multirow{5}*{\includegraphics[width=0.78\linewidth]{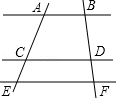}}
     \end{minipage} 
     &
     \begin{minipage}[t]{0.32\textwidth}  
     \texttt{Line A B}\\
     \texttt{Line C D}\\
     \texttt{Line E F}\\
     \texttt{Line E C A}\\
     \texttt{Line B D F}\\
     \end{minipage} 
     \\
    \bottomrule
    \end{tabular}
    }  
    \caption{Four examples of our formalized diagram-caption pairs containing two relationships among points in geometry images.
}
    \label{tab:caption}
\end{table*}

In Table~\ref{tab:caption}, we provide additional examples of descriptions that delineate the collinear and concyclic relationships in geometric images at the granularity of points. It is noteworthy that we adhered to strict grammatical and standardization criteria during the annotation process. Specifically, for collinear relationships, the term \texttt{Line} denotes the relationship, and the order of the points is listed from left to right. For concyclic relationships, the symbol \texttt{$\backslash\backslash$odot} signifies the center of the circle, \texttt{lieson} indicates the points on the circumference, and the points are listed in a clockwise order.

\vspace{-2pt}
\section{More Evaluations}
\label{app:sec:study}
\vspace{-2pt}

Inspired by the Choice metric proposed by~\citet{pgpsnet/zhang2023multi}, we introduce an accuracy metric for GeoX to ensure \textbf{complete fairness} when comparing with solver-free methods like G-LLaVA~\citep{g-llava/gao2023g}.
Specifically, we observe that  even if errors occur in the solving process, solver-free methods can still provide an answer (by randomly selecting one from four options), whereas our solver-based approach considers any process errors as incorrect results.
To this end, in comparison with solver-free methods as shown in~\cref{table:gllava_comparison}, we define GeoX's accuracy  by assuming that, when the solution process encounters errors, the model's performance is equivalent to randomly selecting from four possible options.
%
We also evaluate our method against solver-free approaches on GeoQA~\citep{geoqa_ngs/chen2021geoqa}. As shown in \cref{table:gllava_comparison}, our method outperforms the current state-of-the-art solver-free methods in Top-1 accuracy.

\begin{table*}[htbp]
\vspace{-10pt}
    \caption{
        Comparison with solver-free geometry specialists on GeoQA. We directly report results using Top-1 accuracy.
        }
    \centering
    \begin{minipage}[htbp]{0.6\textwidth}
        \flushright
        \makeatletter\def\@captype{table}
        \resizebox{\textwidth}{!}{
        \begin{tabular}{ccc}
        \toprule
        Methods & Base LLM & Accuracy \\ 
        \hline
        Math-LLaVA~\citep{math-llava/shi2024math} & Vicuna-1.5-13B & 48.1 \\
        G-LLaVA~\citep{g-llava/gao2023g} & LLaMA-2-7B & 64.2 \\
        MAVIS~\citep{mavis/zhang2024mavis} & MAmmoTH-2-7B & 66.7 \\
        EAGLE~\citep{eagle/li2024eagle} & Vicuna-1.5-7B & 67.1 \\
        \rowcolor[gray]{0.9} 
        GeoX (Ours) & Geo-LLM-7B& \textbf{67.4} \\
        \bottomrule
        \end{tabular}
        }
        \label{table:gllava_comparison}
    \end{minipage}    
\end{table*}






\vspace{-2pt}
\section{Data Acquisition for Geometric Corpus}
\label{app:sec:corpus}
\vspace{-2pt}

\paragraph{Data Sources.} We detail the geometric corpus collections used to train Geo-LLM, sourced from a variety of publicly available geometric datasets, including GeoQA~\citep{geoqa_ngs/chen2021geoqa}, GeoQA+\citep{geoqa_ngs/chen2021geoqa}, UniGeo\citep{unigeo_geoformer/chen2022unigeo}, PGDP5K~\citep{pgdp5k/hao2022pgdp5k}, PGPS9K~\citep{pgpsnet/zhang2023multi}, Geometry3K~\citep{inter-gps/lu2021inter}, and G-LLaVA~\citep{g-llava/gao2023g}. 

\begin{itemize}
    \item \paragraph{GeoQA~\citep{geoqa_ngs/chen2021geoqa}} comprises 4,998 real-world geometry problems sourced from Chinese middle school exams, each annotated with detailed solution processes and human performance metrics. The dataset is organized into three primary categories: angle, length, and other geometric calculations, and is divided into training, validation, and test sets at a ratio of 7.0:1.5:1.5.
    
    \item \paragraph{Geometry3K~\citep{inter-gps/lu2021inter}} provides 3,002 detailed geometry problems derived from high school textbooks, divided into training, validation, and test sets in a 0.7:0.1:0.2 ratio. Geometry3K expands on previous datasets~\citep{geos/seo2015solving} by including irregular quadrilaterals, polygons, and additional unknown variables and operator types. Moreover, less than 1\% of Geometry3K problems can be solved without diagrams, making it more challenging.

    \item \paragraph{GeoQA+~\citep{dpe-gps/cao2022augmented}} enhances the original GeoQA~\citep{geoqa_ngs/chen2021geoqa} by adding 2,518 newly annotated geometric problems, increasing the total to 7,528 problems with 6,027 dedicated for training. This expanded dataset introduces more complex problems, including area calculations, and raises the difficulty with 27 knowledge points and an average of 2.61 solving steps per problem.
    
    \item \paragraph{UniGeo~\citep{unigeo_geoformer/chen2022unigeo}} introduces a comprehensive geometry dataset encompassing both calculation and proof problems, including 9,543 proving problems sourced from educational websites and 4,998 calculation problems from GeoQA~\citep{geoqa_ngs/chen2021geoqa}. The proof problems are categorized into five sub-tasks (parallel, triangle, quadrangle, congruent, and similarity) with detailed reasoning and expressions. To facilitate unified problem-solving, both proofs and solutions are reformulated into sequence formats, aligning the proving steps with calculation protocols.
    
    \item \paragraph{PGDP5K~\citep{pgdp5k/hao2022pgdp5k}} contains a total of 5,000 images, divided into training, validation, and test sets with a 0.7:0.1:0.2 split. It encompasses 16 geometric shapes, 5 positional relations, 16 symbol types, and 6 text types. PGDP5K provides detailed annotations, including geometric primitives, symbols, text types, and their relationships.

    \item \paragraph{PGPS9K~\citep{pgpsnet/zhang2023multi}} consists of 9,022 geometry problems paired with 4,000 unique diagrams, covering 30 problem types from grades 6-12 mathematics curricula. It is split into training and test sets, with 8,433 samples for training and 589 for testing. PGPS9K includes detailed annotations for diagrams and solution programs.

    \item \paragraph{G-LLaVA~\citep{g-llava/gao2023g}} is a large-scale multi-modal geometry dataset consisting of over 110k question-answer (QA) pairs, divided into an alignment dataset to provide foundational geometric knowledge and an instruction-tuning dataset to improve the model's problem-solving abilities. This dataset is created with the assistance of GPT-API~\citep{chatgpt/ouyang2022training} using various strategies, including equation solving, value scaling, and sentence paraphrasing.

\end{itemize}

\paragraph{Data Collection and Filtering.} To meet the demands of pre-training for Geo-LLM, we build up a specialized filtering and pre-processing pipeline to construct the geometric corpus. Initially, we extract the data only from the training portions from the existing geometric datasets to prevent label leakage. Besides, we use a free Translate-API to convert Chinese data into English, ensuring language consistency. For each data entry, we remove content unrelated to geometric problems, such as annotation IDs, dates~\citep{inter-gps/lu2021inter}, and sources~\citep{pgpsnet/zhang2023multi}. Ultimately, we achieve a collection of 100 million tokens of data.



\vspace{-2pt}
\section{Additional Details}
\label{app:sec:implementation}
\vspace{-2pt}

\paragraph{Prompts for MLLMs.}
In \cref{tab:prompt_for_mllm}, we provide examples of how to prompt multimodal large models to reason on geometric problems across two different evaluation modes.
Each evaluation mode consists of several components: System Prompt, Diagram, Question, and optionally, Choices. The System Prompt specifies the type of problem the model is required to solve and the expected output format. The Diagram corresponds to the relevant image, while the Question and Choices are presented in the text.
The key difference between the Choice and Completion modes is that Completion requires the model to provide answers directly, while Choice only involves selecting from predefined options.

\paragraph{Evaluation Versions for Generalists.} In \cref{tab:lmm_generating_params}, we present the model / API versions utilized for the evaluation of generalists reported in \cref{exp:geoqa,table:unigeo,table:3k9k,table:mathvista}. These include MLLMs such as mPLUG-Owl2~\citep{mplug-owl2/ye2023mplug}, Qwen-VL~\citep{qwen-vl/bai2023qwen}, LLaVA-v1.5~\citep{llava/liu2024visual}, GPT-4V~\citep{openai2023gpt4v}, and GPT-4o~\citep{openai2024gpt4o}.

\paragraph{Implementation Details.} After unified formal vision-language pre-training, we fine-tuned GeoX on each dataset to achieve better performance. The hyperparameters required for end-to-end visual instruction tuning are shown in \cref{tab:appendix:details}.

\begin{table*}[h]
    \centering
    \small
    \renewcommand\tabcolsep{4pt} 
    \resizebox{\textwidth}{!}{  
    \begin{tabular}{c|p{12cm}}  
     \toprule
     \textbf{Eval Mode} & \textbf{Prompt}\\
     \midrule
    \textit{Choice}  & 
     \begin{minipage}[t]{\textwidth}  
     \textbf{System Prompt:} You are an intelligent robot expert at solving geometry problems. Please ans-\\wer the Question based on the image. You should provide the reasoning process, and then you\\ must give the correct choice in the end based on your reasoning in the following form: \\ The answer is (A), (B), (C) or (D).\\
    \textbf{Diagram}: The Diagram is \texttt{<img>{image\_id.png}</img>}\\
     \textbf{Question:} As shown in the figure, in triangle A B C , it is known that angle A = 80.0 , angle B \\ = 60.0 , D E parallel B C , then the size of angle C E D is (). \\
    \textbf{Choices:} (A) 40.0 (B) 60.0 (C) 120.0 (D) 140.0
     \end{minipage}
     \\
     \midrule
    \textit{Completion}  & 
     \begin{minipage}[t]{\textwidth}  
     \textbf{System Prompt:} You are an intelligent robot expert at solving geometry problems. Please ans-\\wer  the Question based on the image. You should provide the reasoning process, and then you \\ must give the correct answer in the end based on your reasoning in the following form: \\e.g., The answer is [12.1]. \\
     \textbf{Diagram}: The Diagram is \texttt{<img>{image\_id.png}</img>}\\
     \textbf{Question:} 
     Line m is the perpendicular bisector of XZ, WZ = 14.9. Find WX.
    \end{minipage}
    \\
    \bottomrule
    \end{tabular}
    }  
    \caption{The prompts used for Choice and Completion modes in Multi-modal Large Language Models (MLLMs). To guide MLLMs in reasoning on geometric tasks, we adopt two evaluation modes like~\citet{lans/zhang2023lans}: Choice and Completion.
}
    \label{tab:prompt_for_mllm}
\end{table*}

\begin{table}[th!]
    \small
    \centering
    \begin{tabular}{lc}
    \toprule
    \textbf{Model Name} & \textbf{Model / API Version} \\
    \midrule
    mPLUG-Owl2~\citep{mplug-owl2/ye2023mplug} & mplug-owl2-llama2-7b \\ 
    \midrule
    LLaVA-v1.5~\citep{llava/liu2024visual} & llava-v1.5-13b-hf   \\
    \midrule
    Qwen-VL~\citep{qwen-vl/bai2023qwen} & Qwen-VL-Chat \\
    \midrule
    GPT-4V~\citep{openai2023gpt4v} & gpt-4-vision-preview \\
    \midrule
    GPT-4o~\citep{openai2024gpt4o} & gpt-4o-2024-05-13 \\ 
    \bottomrule
    \end{tabular}
    \caption{Model / API versions used for evaluation across different MLLMs.}
    \label{tab:lmm_generating_params}
\end{table}

\begin{table}[H]
    \centering
    \small
    \resizebox{0.7\columnwidth}{!}{%
        \begin{tabular}{@{}lcccc@{}}
        \toprule
        Instruction Tuning &  GeoQA & UniGeo  & PGPS9K & Geometry3K
        \\ \toprule
        Training Batch Size & \multicolumn{4}{c}{64} \\
        Scheduler & \multicolumn{4}{c}{Cosine Annealing} \\
        Optimizer & \multicolumn{4}{c}{AdamW} \\
        \cline{1-1}
        Warmup Ratio &  0.05 &  0.05 &  0.05 &  0.03 \\
        Epochs & 100 & 80 & 45 & 30 \\
        Learning Rate & 3e-5 & 3e-5 & 6e-5 & 2e-5 \\
        Evaluation Steps & 200 & 400 & 200& 200\\
        \bottomrule
        \end{tabular}%
    }
    \caption{Hyperparameters for end-to-end visual instruction tuning. We finetune these models on 4 A100 (80GB) GPUs, respectively.}
    \label{tab:appendix:details}
\end{table}



\section{Further Discussion}
\label{app:sec:gpt4v_formalprogram}

\begin{figure}[H]
    \vspace{-4pt}
    \centering
    \includegraphics[width=0.85\linewidth]{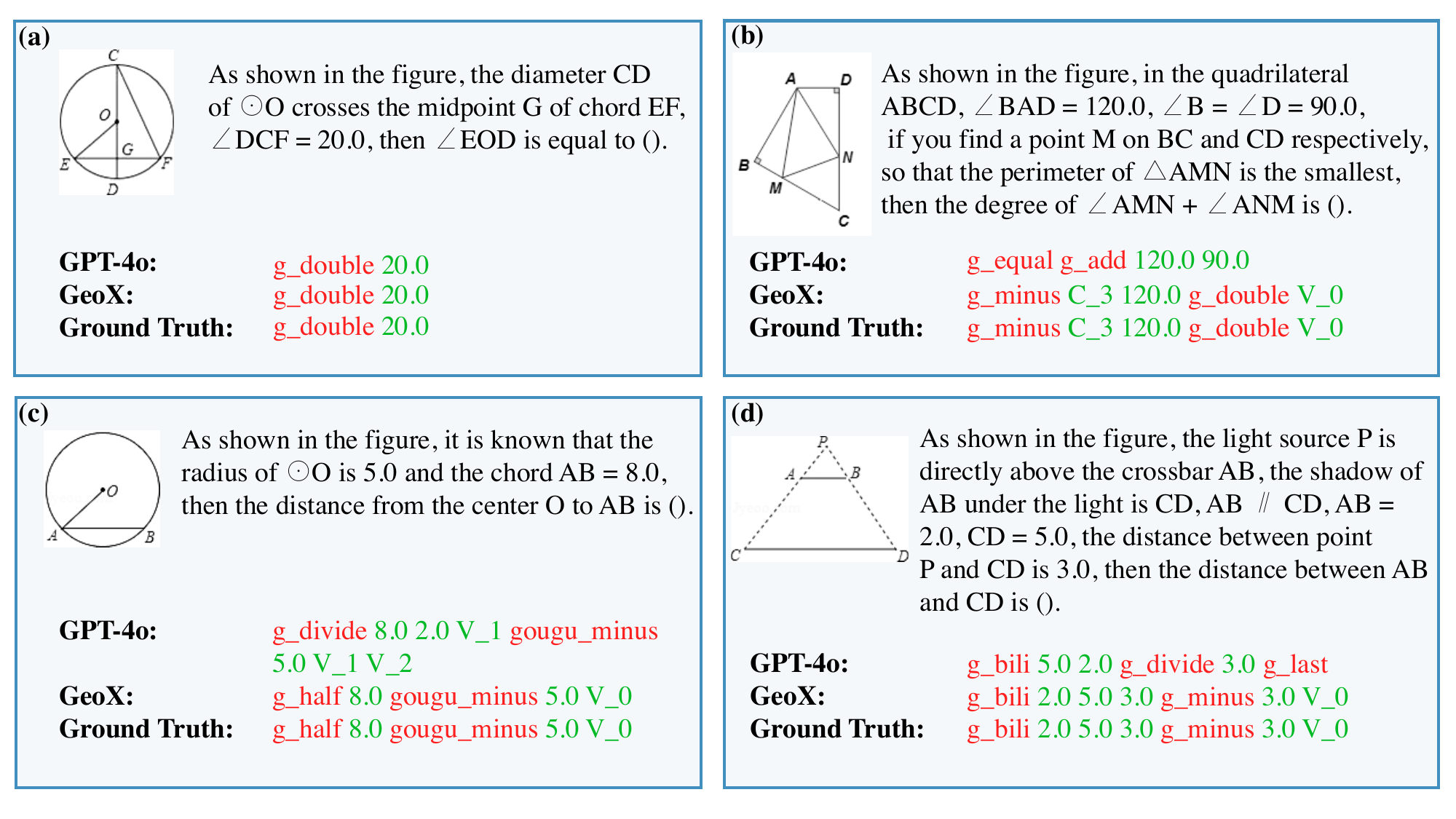}
    \vspace{-6pt}
    \caption{
    Comparison of GPT-4o and GeoX in predicting formalized programs for solving complex geometric problems. 
    }
    \label{fig:4o_formal}
    \vspace{-4pt}
\end{figure}

\paragraph{Analysis of advanced MLLMs' Ability in Formal Programs Generation.} 
As shown in \cref{table:mathvista}, GPT-4o~\citep{openai2024gpt4o} demonstrates the highest accuracy on MathVista-GEO. 
In this section, we delve deeper into the few-shot learning ability of GPT-4o's in generating formalized program sequences, which are then sent to the GPS solver for solving~\citep{unigeo_geoformer/chen2022unigeo}.
Specifically, we apply 2-shot in-context learning, providing GPT-4o~\citep{openai2024gpt4o} with two examples of formal problem-solving, along with the complete set of operation functions.
Then, GPT-4o is tasked with predicting the corresponding solving program when presented with new problems and geometric images.
As shown in \cref{fig:4o_formal}, GPT-4o~\citep{openai2024gpt4o} is capable of predicting simple geometric programs, but for more complex problems, it exhibits issues such as predicting only the operation without the variable (e.g., \texttt{g\_equal} in b), incorrect variables (e.g., \texttt{gougu\_minus 5.0 V\_1 V\_2} vs \texttt{gougu\_minus 5.0 V\_0} in c), or wrong operations (e.g., \texttt{g\_equal} vs \texttt{g\_minus} in d). In contrast, GeoX can predict the correct solution in these complex and diverse cases.


\end{document}